# Modeling Vehicle-Type-Specific Pedestrian Crash Avoidance Behavior in Safety-Critical Interactions Using Smooth-Mamba Deep Reinforcement Learning

Qingwen Pu[a], Kun Xie[a,*], Hong Yang[b], Di Yang[c], Junqing Wang[b]

[a] *Transportation Informatics Lab, Department of Civil and Environmental Engineering, Old Dominion University, Norfolk, VA 23529, United States*
[b] *Department of Electrical and Computer Engineering, Old Dominion University, Norfolk, VA, United States*
[c] *Department of Transportation and Urban Infrastructure Studies, SMARTER Center, Morgan State University, 1700 E Cold Spring Ln, Baltimore, MD 21251, USA*

[*] *Corresponding Author; Email: kxie@odu.edu*

## Abstract

As automated vehicles (AVs) increasingly share roadways with human-driven vehicles (HDVs), understanding how pedestrians respond to different vehicle types in safety-critical interactions is essential for the safe deployment of automated driving technologies. This study extracts safety-critical pedestrian–vehicle interactions from the Argoverse 2 dataset to capture real-world crash avoidance behaviors in encounters involving AVs and HDVs. To model vehicle-type-specific pedestrian crash avoidance behavior, we develop a Smooth-Mamba Deep Deterministic Policy Gradient framework, termed SMamba-DDPG, which integrates smooth action constraints with efficient temporal representation learning. To quantify pedestrian behavioral differences, the framework trains separate crash avoidance policies for pedestrian interactions with AVs and HDVs. Results show that SMamba-DDPG outperforms baseline reinforcement learning and supervised learning models in reproducing pedestrian crash avoidance behaviors. Reconstructed trajectories demonstrate strong behavioral realism, accurately reproducing crash avoidance kinematics in both AV and HDV scenarios. Reaction time analysis shows that the model captures human-like response delays and reveals that pedestrians respond more quickly to AVs than to HDVs. Counterfactual analysis further indicates that pedestrians adopt lower crossing speeds when interacting with AVs. Large-scale safety analysis of model-generated data revealed that pedestrian-AV interactions consistently yielded lower conflict rates and higher pedestrian yielding rates compared to pedestrian-HDV interactions. The findings highlight the importance of incorporating vehicle-type-specific pedestrian behavioral models for safer automated driving system design and more realistic traffic simulations in mixed-traffic environments.

## 1. Introduction

As automated vehicles (AVs) become part of urban traffic, they create mixed environments shared with human-driven vehicles (HDVs). A key research question arises (Lanzaro *et al.* 2023): how do pedestrians respond differently to AVs versus HDVs? Unlike HDV drivers, AVs lack the capacity for eye contact, gestures, or other social cues that pedestrians rely on to negotiate right-of-way, potentially altering pedestrian trust, risk perception, and avoidance behavior (Cui and Oca 2025). Understanding these behavioral differences is essential for reducing collision risk at urban intersections, where pedestrian decisions directly influence vehicle-pedestrian interaction dynamics (Anish *et al.* 2025).

Existing studies primarily examine AV collision avoidance systems such as sensor fusion, path planning, and emergency braking (Tan *et al.* 2024, Cui *et al.* 2025). This vehicle-centric approach overlooks how AV maneuvers influence pedestrian decisions. Traditional car-following and lane-changing models capture normal driving conditions but fail to represent near-collision events where pedestrians must rapidly respond to approaching vehicles, and they neglect how AV actions affect surrounding traffic participants (Aria *et al.* 2016, Giummarra *et al.* 2021). Pedestrian responses depend on cognitive processes such as vehicle-type recognition, motion predictability, and perceived intent (Dommes and Cavallo 2011). Unfamiliarity with AV behavior patterns may induce distinct avoidance strategies (Prédhumeau *et al.* 2021). However, empirical evidence quantifying these differences remains limited (Aittoniemi 2022). Most frameworks aggregate AV and HDV interactions without differentiation, limiting the ability to predict pedestrian behavior in mixed-traffic environments and compromising AV safety system effectiveness (Song and Ding 2023, Pu *et al.* 2026a).

Near-miss scenarios offer unique insights into pedestrian decision-making under extreme time pressure (Tinsley *et al.* 2012). These safety-critical interactions expose the temporal dynamics of risk perception and evasive action that remain hidden during routine crossings (Raiyn and Weidl 2024). Because such real-world cases are rare (Pu *et al.* 2025b), this study uses the Argoverse 2 Motion Forecasting Dataset (Wilson *et al.* 2023), which includes 250,000 high-resolution urban driving scenarios involving AVs, HDVs, and pedestrians. This comprehensive dataset enables systematic extraction and analysis of safety-critical pedestrian–vehicle interactions under mixed-traffic conditions (Li *et al.* 2024).

Traditional statistical approaches and rule-based models struggle to capture the sequential dynamics of pedestrian avoidance trajectories (Korbmacher and Tordeux 2022). Deep reinforcement learning (DRL) addresses this limitation by learning optimal policies directly from interaction data (Lei *et al.* 2020). However, standard DRL architectures struggle with rare-event learning in safety-critical interactions and exhibit training instability during abrupt state transitions typical of emergency maneuvers (Guo *et al.* 2023). When integrating temporal models with DRL for policy learning, conventional architectures such as LSTM-based DRL and Transformer-based DRL face similar stability challenges (Guo *et al.* 2024). Recent advances in structured state-space models integrated with DRL, particularly Mamba-DRL, demonstrate superior capability in filtering noise and capturing long-range dependencies (Lu *et al.* 2025). However, Mamba exhibits substantial outliers in gate and output projections, amplified by parallel scan operations, leading to distributional imbalances that hinder parameter compression and computational efficiency (Xu *et al.* 2025).

To overcome these challenges, this study models pedestrian crash avoidance behaviors to examine how pedestrians respond differently to AVs versus HDVs in safety-critical interactions. We propose the Smooth-Mamba Deep Deterministic Policy Gradient (SMamba-DDPG) framework, which integrates structured state-space modeling to capture temporal dependencies and incorporates smoothing mechanisms to stabilize learning in rare events. The integration of the Argoverse 2 dataset provides high-resolution trajectory data across diverse urban driving scenarios. Separate pedestrian-centered policy models trained for each vehicle type enable counterfactual analysis and safety assessment, enhancing pedestrian behavior modeling and AV safety system design in mixed-traffic environments. The main contributions of this study include:

(1) This study is among the first to quantitatively model and compare pedestrian crash avoidance behaviors in safety-critical interactions with AVs versus HDVs using a DRL framework.
(2) It incorporates Mamba state-space models with smoothing mechanisms to stabilize learning and address distributional imbalances during emergency maneuvers.
(3) It establishes a comprehensive validation and analysis framework including reaction-time analysis, counterfactual experiments, large-scale conflict-risk evaluation, and yielding behavior assessment under mixed-traffic conditions.

## 2. Literature Review

### 2.1. Distinct pedestrian responses to AVs and HDVs

Current AV research primarily focuses on enhancing vehicles' self-protective safety responses (Furlan *et al.* 2020). For instance, studies analyze emergency braking algorithms when AVs detect pedestrians (Cicchino 2022), vehicle-to-vehicle collision avoidance in mixed traffic (Ma *et al.* 2023), and sensor fusion methods for environmental perception (Wang *et al.* 2023). In parallel, studies on HDVs have examined driver reaction times and braking patterns in pedestrian scenarios (Li *et al.* 2025a). However, a critical gap remains in understanding how pedestrians modify their behavior during safety-critical interactions with vehicles (Rezwana and Lownes 2024).

Previous research has documented pedestrian behavior changes in AV interactions (Mahadevan *et al.* 2019). Observational studies indicate that pedestrians often hesitate to cross because they cannot clearly interpret the intentions of AVs (Zhao *et al.* 2022). Field experiments also reveal that pedestrians exhibit greater caution when interacting with AVs, often waiting for explicit cues such as vehicle deceleration before crossing (Tian *et al.* 2023). A VR study found that pedestrians trusted AVs more when vehicles drove defensively and when crosswalks had traffic signals (Jayaraman *et al.* 2019). However, these studies focus on AVs in isolation and lack controlled experiments that compare pedestrian responses to AVs and HDVs (Lanzaro *et al.* 2023).

Although safety-critical interactions are important for understanding real emergency reactions, most studies still focus on normal or early-stage crossing behavior (Kavta *et al.* 2025). Few works analyze pedestrian behavior when collision is imminent or compare AV and HDV cases under such conditions. For example, Hübner *et al.* (2025) examined pedestrian hesitation and communication during crossing decisions in vehicle–pedestrian interactions. Mirzabagheri *et al.* (2025) studied pedestrian intention prediction in risky situations, but these cases mainly described approach phases rather than imminent collision events. Similarly, simulator-based studies (Papadopoulos *et al.* 2024) focus on decision-making at the start of crossing rather than on how pedestrians perform emergency avoidance when a collision becomes imminent. Understanding pedestrian avoidance patterns when interacting with different vehicle types in safety-critical interactions is essential for developing effective AV safety systems and realistic mixed-traffic simulations (Mirzabagheri *et al.* 2025).

Three critical gaps limit understanding of vehicle-type-specific pedestrian responses. First, current studies focus mainly on vehicle safety responses rather than on pedestrians' behavioral changes during interactions (Rezwana and Lownes 2024). Second, existing research lacks comparisons of pedestrian responses to AVs versus HDVs, making it unclear how each vehicle type affects pedestrian behavior (Lanzaro *et al.* 2023). Third, most studies address typical crossing interactions rather than emergency avoidance behaviors in safety-critical interactions (Pu *et al.* 2026b).

Addressing these gaps requires a framework that (1) models pedestrian behavior separately for AV and HDV interactions under identical conditions, (2) focuses on safety-critical interactions where differences in speed, acceleration, and collision avoidance strategies are most pronounced and safety risk is highest, and (3) enables counterfactual analysis to quantify behavioral differences between vehicle types and to conduct large-scale safety analysis using model-generated interactions. Such an approach would reveal whether pedestrians exhibit distinct avoidance patterns based on vehicle type and quantify these differences in terms of measurable safety metrics.

### 2.2. Deep reinforcement learning in pedestrian avoidance behavior modeling

Traditional statistical and rule-based models fail to capture the dynamic and adaptive characteristics of pedestrian avoidance trajectories (Korbmacher and Tordeux 2022). Social force models describe pedestrian movement through attractive and repulsive forces but oversimplify decision-making dynamics (Chen *et al.* 2018). Velocity obstacle methods predict collision-free paths using relative velocities (Zhang *et al.* 2017). However, these approaches fail to model the sequential dependencies in continuous avoidance maneuvers (Yang *et al.* 2024). During safety-critical interactions, pedestrians continuously adjust their speed and direction in response to real-time vehicle movements (Pu *et al.* 2026c); however, rule-based models cannot capture such adaptive behaviors (Kelley *et al.* 2025).

Trajectory prediction methods forecast future pedestrian paths based on observed trajectories and surrounding dynamics. Early approaches such as LSTM-based models learn short-term temporal dependencies in pedestrian motion sequences (Quan *et al.* 2021), while Transformer-based models capture long-range spatial–temporal interactions for more accurate trajectory forecasting (Lorenzo *et al.* 2021, Chen *et al.* 2023). Although effective in regular scenarios, these methods act as passive predictors that estimate trajectories without modeling pedestrians' decision-making mechanisms (Li *et al.* 2025b). This limitation becomes critical in safety-critical interactions, where understanding the reasons behind pedestrians' avoidance strategies is essential (Cheng *et al.* 2024).

Deep reinforcement learning (DRL) addresses these limitations by modeling pedestrians as active decision-makers. Unlike trajectory prediction, DRL captures the sequential decision-making process. In this process, an agent learns to observe vehicle states, select appropriate actions, and receive safety-related feedback (Wu *et al.* 2024). Recent work has demonstrated success in learning human-like avoidance strategies (Emuna *et al.* 2020, Yavas *et al.* 2023). However, standard DRL architectures struggle to maintain temporal continuity in modeling pedestrian–vehicle interactions. Conventional frameworks such as DDPG and TD3 treat each state independently, limiting their capacity to represent evolving motion patterns (Shen 2024, Pu *et al.* 2026d)ss. Consequently, these models often fail to produce adaptive behaviors in safety-critical interactions (Mirzabagheri *et al.* 2025).

To enhance policy stability and improve learning efficiency in dynamic environments, recent studies have incorporated temporal models into DRL frameworks. LSTM-based DRL captures temporal dependencies through recurrent connections in actor-critic networks (Tseng and Ferng 2025), whereas Transformer-based DRL employs self-attention to model long-range dependencies in state sequences (Guo *et al.* 2024). However, both architectures struggle with sudden motion changes in emergency avoidance. Abrupt accelerations, sharp turns, and rapid stops create gradient instability, causing convergence difficulties (Attia and Koren 2021). LSTM-DRL suffers from vanishing gradients in extended sequences (Paul *et al.* 2024). Transformer-DRL faces computational challenges as attention complexity scales quadratically with sequence length (Liang *et al.* 2023).

Structured state-space models combined with DRL effectively filter noise and capture long-range dependencies. The Mamba architecture enhances the ability through selective state-space modeling and efficient parallel processing (Gu and Dao 2024). Within DRL frameworks, Mamba emphasizes key decision points, making it suitable for safety-critical interactions (Lu *et al.* 2025). However, Mamba may produce outliers in gate and output projections when distinguishing between critical and non-critical states (Xu *et al.* 2025). In safety-critical interactions, the contrast between routine crossing and emergency evasion creates extreme activation values. These imbalances hinder parameter compression and computational efficiency, as outlier values exceed the typical bit-precision ranges that standard quantization can handle (Kim *et al.* 2025).

Smoothing normalizes channel-wise variance and suppresses activation spikes (Xu *et al.* 2025), stabilizing gradient flow during abrupt state transitions. Unlike global normalization, it operates locally within gating and projection paths, preserving the model's sensitivity to safety-critical events while reducing distributional imbalances. Building on this mechanism, this study develops the SMamba-DDPG framework, which enables learning from rare safety-critical interactions while preserving temporal coherence. This integration is essential for distinguishing pedestrian behavioral differences when interacting with AVs versus HDVs.

## 3. Data

The Argoverse 2 Motion Forecasting Dataset (Wilson *et al.* 2023) is employed to analyze pedestrian crash avoidance behaviors in safety-critical interactions involving AVs and HDVs. Instead of focusing solely on Pedestrian-AV scenarios, this study systematically extracts and compares critical Pedestrian-AV and Pedestrian-HDV interactions. The dataset comprises 250,000 curated driving scenarios recorded at 10 Hz over 11-second intervals. These scenarios provide high-resolution 2D centroid and heading information for various road users, captured across six major U.S. cities with over 2,000 km of diverse road infrastructure. Each trajectory track includes detailed attributes such as object ID, position, heading, velocity, and category. Each scenario includes a high-definition map detailing road geometry, lanes, and crosswalks, which supports spatial reasoning for detecting vehicle–pedestrian interactions.

In this dataset, the AV label refers to the Argoverse data-collection ego vehicle. This vehicle is equipped with two LiDAR units, two stereo cameras, and seven ring cameras providing 360° perception. Due to its visible sensor hardware, pedestrians may recognize it as an autonomous or test vehicle, while all other vehicles are classified as HDVs. As illustrated in Figure 1 (a) and (b), the dataset captures bird's-eye views of dense urban intersections where AVs (green), HDVs (blue and orange), and pedestrians (purple) interact, highlighting safety-critical interactions such as turning maneuvers, pedestrian crossings, and multi-agent congestion. Figure 1 (c) shows the AV data collection vehicle described above, which was deployed across the six U.S. cities to construct the dataset.

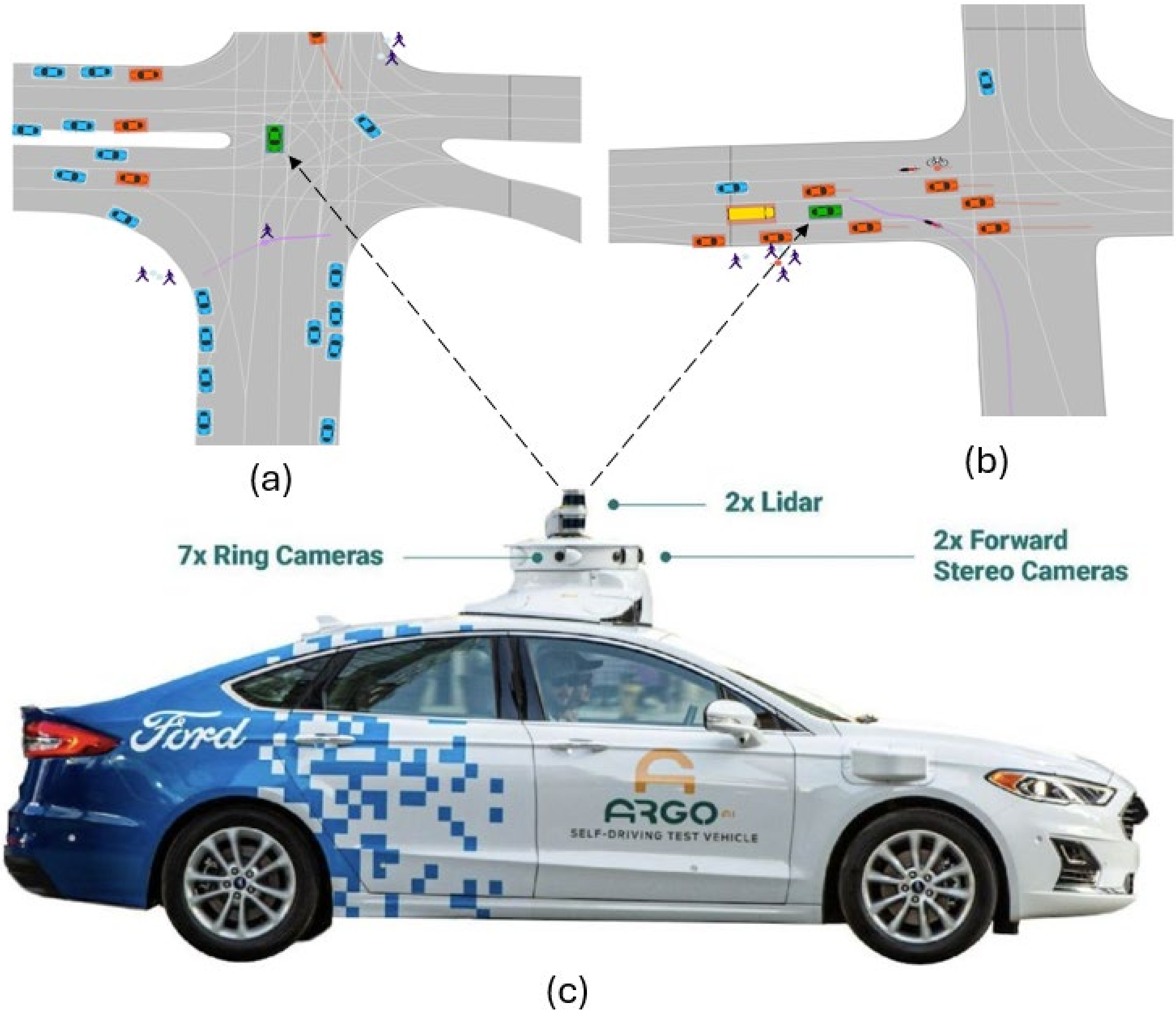

**Figure 1 : (a) and (b) Example urban interaction scenarios with pedestrians and vehicles from the Argoverse 2 dataset; (c) The self-driving vehicle platform used for data collection. (Source: (Wilson *et al.* 2023))**

The Argoverse 2 Motion Forecasting Dataset provides comprehensive trajectory data, capturing detailed geospatial and kinematic parameters essential for in-depth interaction analysis. As summarized in Table 1, each data entry contains key fields that describe the dynamic state and context of objects within a scenario.

**Table 1 Data Dictionary for Argoverse 2 Motion Forecasting Dataset**

| Field Name | Description |
|---|---|
| scenario_id | Unique ID for each 11-second scenario. |
| track_id | Unique ID for each tracked object/actor. |
| object_type | General object classification. |
| object_category | Specific category for object type (dynamic or static). |
| timestep | Discrete time step (10 Hz). |
| position_x | X-coordinate of object centroid (m). |
| position_y | Y-coordinate of object centroid (m). |
| heading | Object orientation (yaw) in radians. |
| velocity_x | X-component of object velocity (m/s). |
| velocity_y | Y-component of object velocity (m/s). |
| observed | Boolean: timestep is in observed history. |
| focal_track_id | track_id of the primary/focal agent in the scenario. |
| city | City of data collection. |
| num_timestamps | Total timesteps in scenario. |

To enable unified modeling of pedestrian-vehicle collision avoidance behaviors, a structured data processing pipeline is applied, as illustrated in Figure 2. Each scenario file from the Argoverse 2 Motion Forecasting Dataset is first parsed to identify and separate the trajectories of AVs and HDVs based on the track_id metadata. Scenarios that lack valid data for either vehicles or pedestrians are excluded from further analysis.

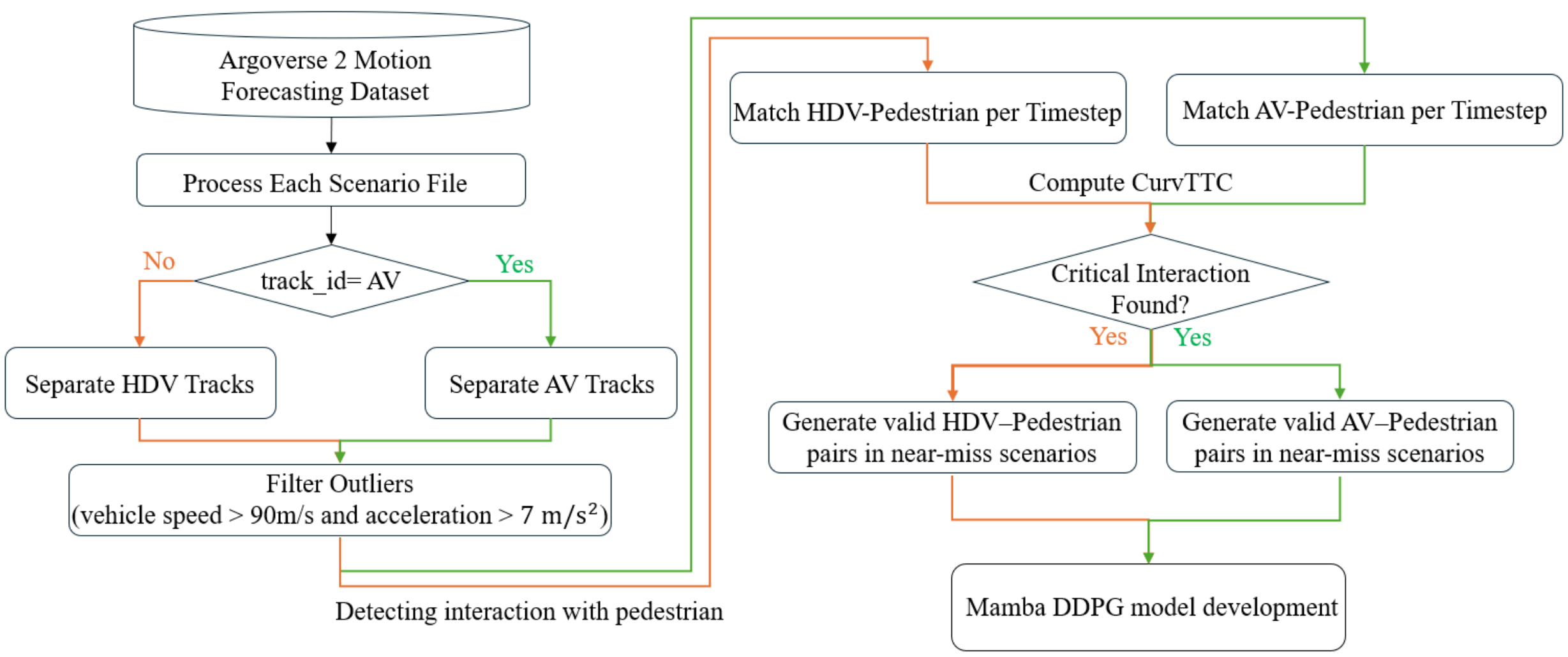

**Figure 2: Data-processing pipeline for extracting Pedestrian-AV and Pedestrian-HDV interactions**

To ensure data quality, records exceeding the maximum speed (90 m/s) or acceleration (7 m/s$^2$) thresholds were treated as outliers and removed (Book 2013). For each remaining case, only Pedestrian-AV and Pedestrian-HDV trajectory segments with true spatial overlap were retained for further analysis.

To identify such interactions, this research computes pairwise Euclidean distances between all pedestrian and vehicle trajectory points. A pedestrian–vehicle pair $(i, j)$is classified as a high-proximity interaction when their trajectory envelopes reach a spatial intersection within a threshold of $d_{\text{thresh}} = 0.1$ m, representing the extreme limit of safety-critical avoidance behaviors. Formally, the interaction set is defined as:

$$\mathcal{I} = \left\{ (i, j) \mid \min_{t_i \in \mathcal{T}_i, t_j \in \mathcal{T}_j} \left\| p_i(t_i) - v_j(t_j) \right\| < d_{\text{thresh}} \right\} \tag{1}$$

where $p_i(t_i)$ denotes the position of pedestrian $i$ at time $t_i$, $v_j(t_j)$ represents the position of vehicle $j$ at time $t_j$, and $T_i, T_j$ are the corresponding time sequences of their trajectories. This formulation ensures that only pairs with spatially overlapping trajectories are retained as potential interactions.

Figure 3 shows representative pedestrian–vehicle interaction cases from six Argoverse 2 cities. Each high-definition map visualization depicts green vehicle and red pedestrian trajectories, highlighting typical conflict situations such as crosswalk incursions, mid-intersection encounters, and multi-lane crossings.

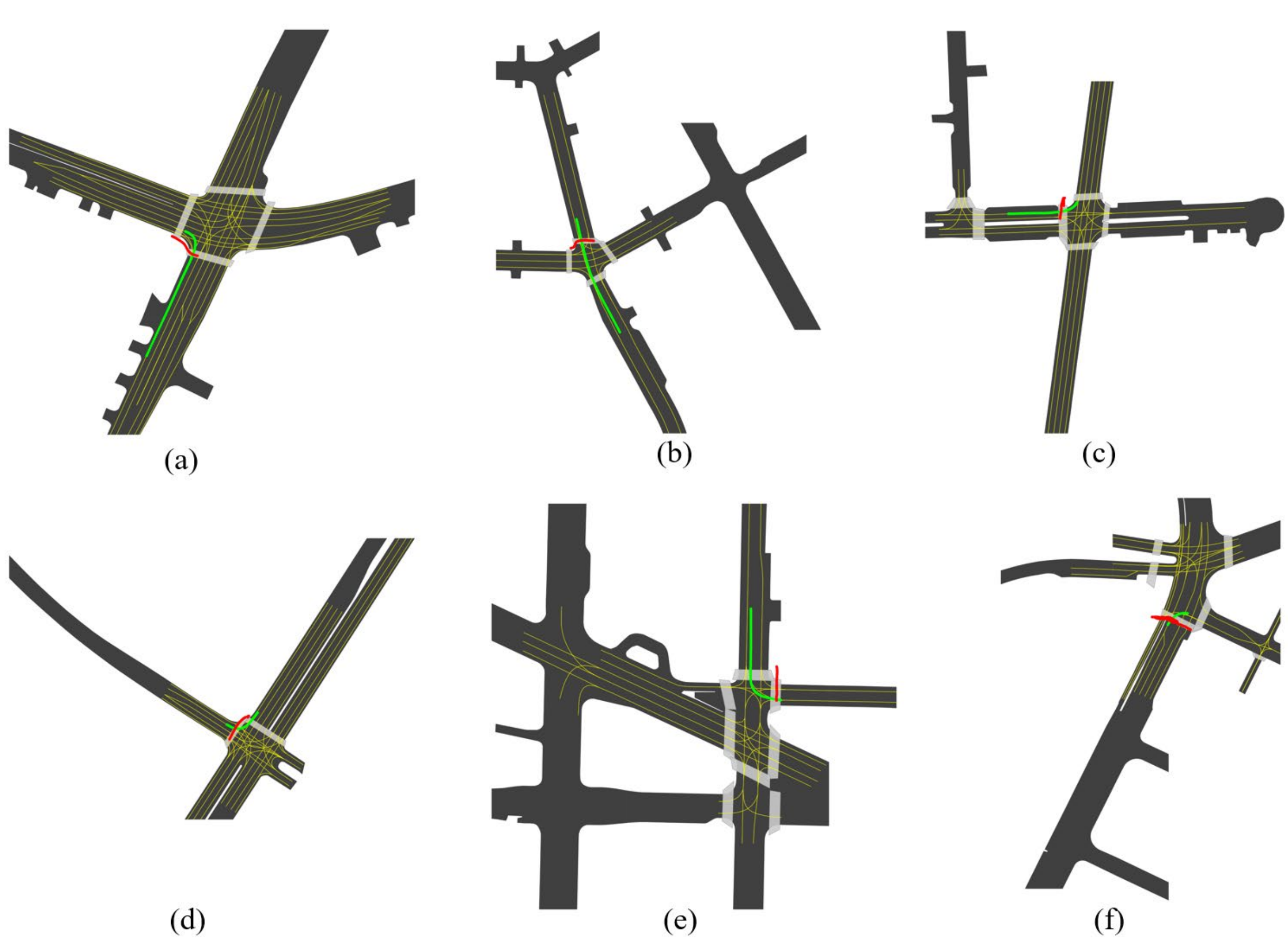

**Figure 3: (a)-(c) Pedestrian-AV interactions; (d)-(f) Pedestrian-HDV interactions**

High-risk vehicle–pedestrian interactions were identified using the Curvilinear Time-to-Collision (CurvTTC) metric (Pu *et al.* 2025a). Traditional surrogate safety indicators such as TTC (Hoffmann and Mortimer 1994), 2D-TTC (Guo *et al.* 2023), and Post-Encroachment Time (Cooper 1984) capture imminent conflicts on straight or simplified trajectories. However, these measures are limited when applied to intersection environments where vehicle paths are curved and pedestrian motion is non-linear (Golchoubian *et al.* 2023). TTC typically assumes straight-line motion and constant velocity, while PET evaluates post-crossing temporal gaps and therefore cannot detect pre-encroachment avoidance behaviors that characterize safety-critical interactions.

CurvTTC overcomes these limitations by explicitly modeling the curvilinear kinematics of turning vehicles and nonlinear pedestrian trajectories. Figure 4 illustrates the complete process of trajectory prediction, collision detection, and CurvTTC computation in a typical urban intersection scenario involving a left-turning vehicle and a crossing pedestrian. The vehicle's velocity is decomposed into longitudinal ($v_{\text{veh,lon}}$) and lateral ($v_{\text{veh,lat}}$) components, while the pedestrian follows a parabolic trajectory ($v_{\text{ped}}$). The red dot marks the predicted collision point, where the agents' projected paths intersect. This setup captures the core concept of CurvTTC—evaluating collision risk based on curved vehicle trajectories and nonlinear pedestrian movements.

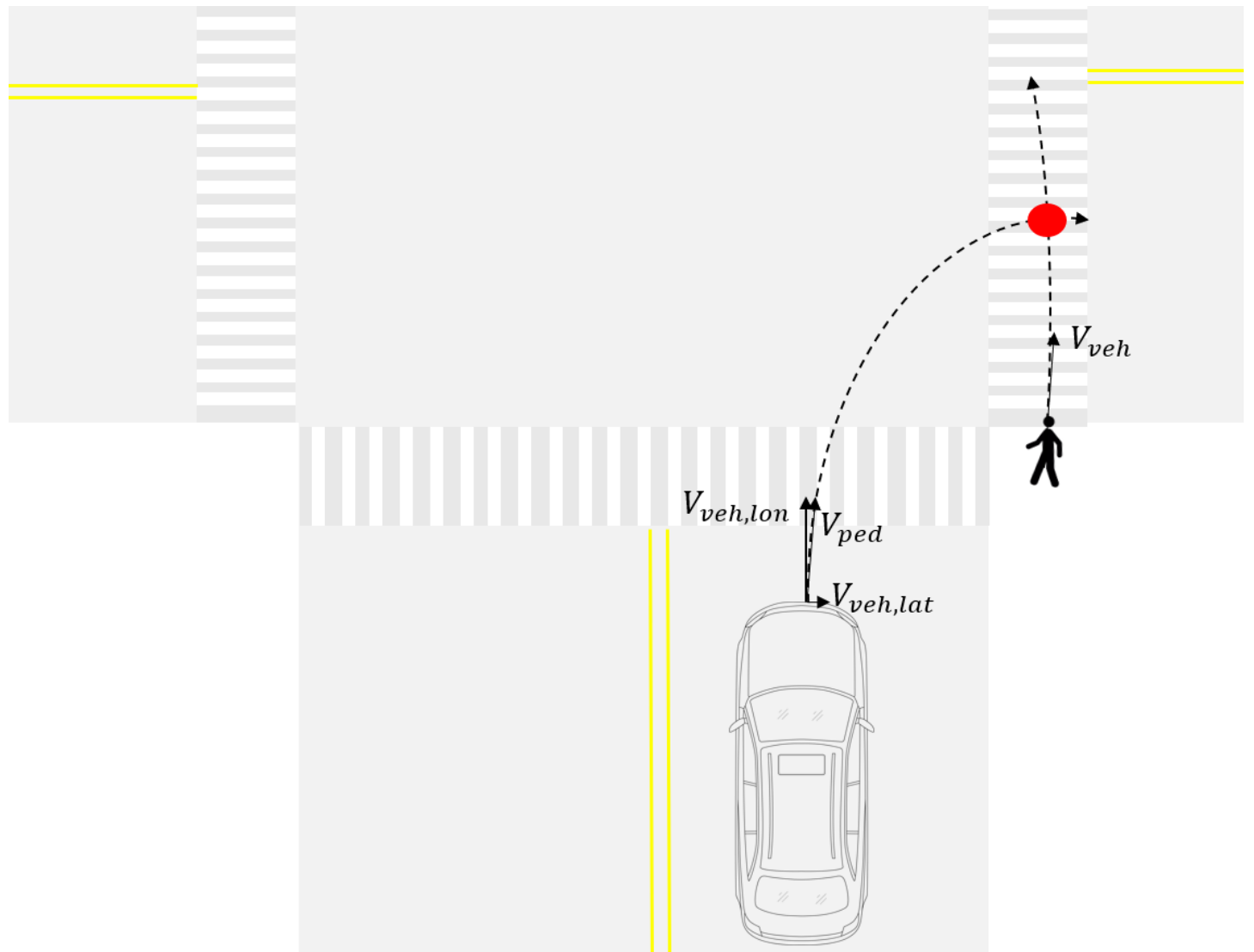

**Figure 4: Conceptual Illustration of CurvTTC in a Vehicle-Pedestrian Interaction**

The algorithm predicts future trajectories of vehicles and pedestrians. Vehicle paths are modeled as circular arcs based on three consecutive positions to capture turning behavior, while pedestrian trajectories are estimated using quadratic interpolation to reflect non-linear movement. At each step, the Euclidean distance between predicted positions is calculated and compared to a collision threshold accounting for both agents' sizes. If a potential collision is identified, the interaction is classified as frontal or lateral based on motion direction and angle. CurvTTC is defined as the time for the vehicle, along its curved path, to reach the projected collision point. Interaction is considered safety-critical if CurvTTC remains below 5 seconds for at least 10 consecutive frames (1 second at 10 Hz).

Using the CurvTTC criterion, 411 Pedestrian-AV and 2,404 Pedestrian-HDV safety-critical interactions were identified, spanning 8,865 and 52,895 time steps, respectively, with 80% of each dataset used for training and 20% for testing. The sample imbalance reflects the real-world scarcity of pedestrian–AV events given the single ego-vehicle design of Argoverse 2. To address this, the two models were trained independently with separate hyperparameter tuning, and evaluation metrics were computed within each group to ensure comparisons reflect behavioral differences rather than sample size effects. The dataset structure is shown in **Table 2**.

**Table 2 Description of key variables related to conflicts**

| Variable | Description |
|---|---|
| Vehicle, Pedestrian | Unique identifiers for (AV or HDV) and pedestrian |
| Time $(t)$ | Timestamp at time step $(t)$, sampled at 10 Hz |
| $X_{\mathrm{AV}}(t), Y_{\mathrm{AV}}(t)$ | AV's lateral and longitudinal positions |
| $V_{\mathrm{AV}}^{x}(t), V_{\mathrm{AV}}^{y}(t)$ | AV's velocity in x- and y-directions |
| $a_{\mathrm{AV}}^{x}(t), a_{\mathrm{AV}}^{y}(t)$ | AV's acceleration in x- and y-directions |
| $X_{\mathrm{ped}}(t), Y_{\mathrm{ped}}(t)$ | Pedestrian's lateral and longitudinal positions |
| $V_{\mathrm{ped}}^{x}(t), V_{\mathrm{ped}}^{y}(t)$ | Pedestrian's velocity in x- and y-directions |
| $a_{\mathrm{ped}}^{x}(t), a_{\mathrm{ped}}^{y}(t)$ | Pedestrian's acceleration in x- and y-directions |
| CurvTTC(t) | Curvilinear time-to-collision between pedestrian and vehicle |

## 4. Methodology

This study investigates differences in pedestrian behavior during safety-critical encounters with AVs and HDVs. Pedestrian-AV interactions involve the Argoverse-2 ego vehicle, which is identifiable by roof-mounted LiDAR and camera sensors. These visible sensors may affect pedestrian perception and crossing decisions. Pedestrian-HDV interactions include encounters with all other vehicles in the dataset. Using the Argoverse 2 Motion Forecasting Dataset, this research extracted two subsets of vehicle–pedestrian interactions: one containing only ego-vehicle encounters and another containing non-ego vehicle encounters. We trained separate pedestrian avoidance policy models on the two subsets to attribute behavioral differences to vehicle type.

### 4.1. SMamba-DDPG framework

Traditional Deep Deterministic Policy Gradient (DDPG) models struggle with three key challenges in safety-critical interactions. First, rare-event learning: safety-critical interactions are infrequent but critical for safety analysis. Second, sequential dependencies: pedestrian avoidance behavior evolves over time based on vehicle movements, but standard DDPG processes states independently. Third, training instability: abrupt state transitions when collision becomes imminent create large gradients that destabilize training. To address these challenges, this study develops the SMamba-DDPG framework, which integrates the Mamba structured state space model (SSM) into DDPG to capture long-range temporal dependencies while incorporating smoothing mechanisms at strategic locations within the Mamba architecture to stabilize learning in rare-event scenarios.

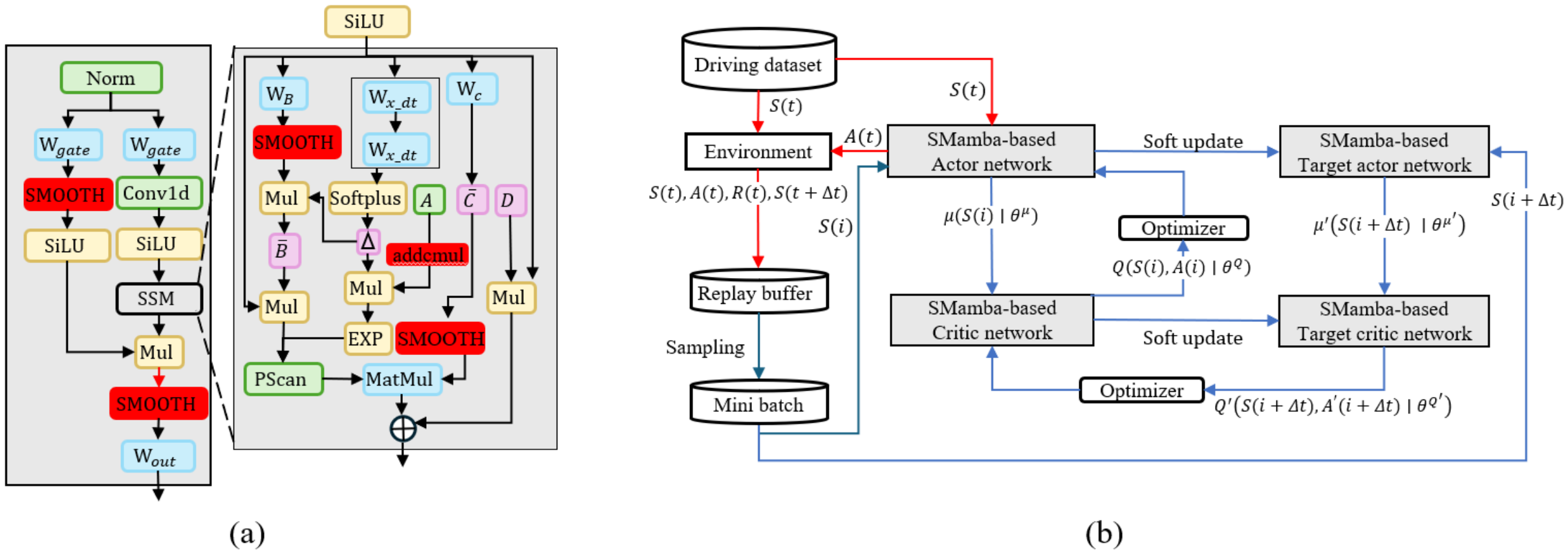

**Figure 5: (a) SMamba block embedded in actor and critic; (b) Overview of SMamba-DDPG structure**

In the proposed framework illustrated in Figure 5, the SMamba-DDPG architecture comprises two main components: a structured state-space module for sequential data processing and a reinforcement learning loop for policy optimization. Figure 5 (a) shows the SMamba block embedded within both the actor and critic networks. This module combines a Selective SSM with a smoothing mechanism, which improves learning stability in environments with frequent and abrupt motion changes. The smoothing module is applied before the Hadamard-based transformation, shown in red in Figure 5 (a). It is inspired by the Smooth-Fused Rotation mechanism from MambaQuant (Xu *et al.* 2025), which was originally designed to reduce instability caused by channel-wise variance in quantized transformer models. In this study, the same approach helps stabilize activations during the learning of rare but critical evasive behaviors. By filtering sharp variations in input features, the model achieves more consistent and reliable learning under dynamic pedestrian-vehicle interactions.

In the output projection path of the SMamba block, the standard SiLU activation is replaced with a smoothed version, termed Smooth-SiLU (S-SiLU). It is defined as:

$$\mathrm{S-SiLU}(x_{\mathrm{act}}, s) = x_{\mathrm{act}} \odot \sigma(s \odot x_{\mathrm{act}}) \tag{2}$$

where $x_{\mathrm{act}}$ denotes the activation input, $s$ is a learnable smoothing vector (broadcast across feature channels), $\sigma(\cdot)$ is the sigmoid function, and $\odot$ represents element-wise multiplication. The smoothing vector $s \in \mathbb{R}^C$ scales the inputs to the nonlinearity. When elements of $s$ are less than 1, they reduce the magnitude of values entering the sigmoid, thereby preventing saturation at extreme activations. This adjustment is particularly beneficial in near-miss pedestrian scenarios, where abrupt behavioral responses may otherwise cause activation spikes and near-binary gating effects.

Within this module, the gating input $x_{\mathrm{gate}}$ is first linearly projected using the matrix $W_{\mathrm{gate}}$ and then passed through the S-SiLU function. The resulting smoothed gate modulates the SSM output $y_{\mathrm{SSM}}$, followed by a linear projection through $W_{\mathrm{out}}$. The overall output is expressed as:

$$y_{\mathrm{out}} = \left[y_{\mathrm{SSM}} \odot \mathrm{S-SiLU}\left(x_{\mathrm{gate}} W'_{\mathrm{gate}}, s_{\mathrm{out}}\right)\right] W'_{\mathrm{out}} \tag{3}$$

Here, $s_{\text{out}}$ is the output-layer smoothing vector, and the primes (') indicate effective weight matrices that have absorbed the smoothing factors, as defined below.

To eliminate additional computational cost during inference, these transformed weights are pre-computed as:

$$W'_{\text{gate}} = W_{\text{gate}} \oslash s_{\text{out}} \tag{4}$$

$$W'_{\text{out}} = s_{\text{out}} \odot W_{\text{out}} \tag{5}$$

where $\oslash$ denotes element-wise division and the operations are broadcast across output channels to ensure dimensional consistency with Eq. (3). This formulation embeds smoothing directly into gating and projection operations, suppressing noise spikes in temporal features and stabilizing learning without additional inference overhead. As a result, the model generates smoother and more realistic pedestrian actions during sudden vehicle maneuvers.

The SSM core governs temporal state evolution via matrix multiplications. In the standard Mamba structure, this involves two parallel branches. In SMamba, smoothing is applied differently to each branch to stabilize hidden-state magnitudes:

Branch 1(C Projection Smoothing):

Output projections through the C-matrix are regularized using a smoothing factor $s_{\text{mm}}$:

$$W'_C = W_C \oslash s_{\text{mm}} \tag{6}$$

This scaling attenuates extreme hidden-state values before propagation, maintaining consistent output ranges.

Branch 2 (Temporal Dynamics Smoothing):

The second branch processes inputs through the B matrix and temporal scaling parameter $\Delta$ using the parallel scan (PScan) operation. The smoothing factor s_mm is split between two sub-paths:

$$W'_B = W_B \oslash s_{\text{mm}} \tag{7}$$

One flows through $\bar{B}$ and is absorbed by $W_B$. The other passes through $\bar{A}$ and is incorporated into the scaling term $\Delta$. Since $\Delta$ is exponentiated, the smoothing is applied via logarithmic adjustment:

$$\text{addcmul}(-\ln(s_{\text{mm}}), \Delta(1), A) = A \cdot \Delta(1) - \ln(s_{mm}) \tag{8}$$

Here, $\Delta(1)$ refers to the value of $\Delta$ at the initial time step, which serves as the anchor for recursive temporal computations. This design ensures that the model can maintain stable activation magnitudes when processing long-horizon sequential inputs.

Figure 5 (b) illustrates the overall SMamba-DDPG framework, which adopts an actor-critic structure. Both the actor and critic networks embed SMamba modules to enhance temporal feature extraction and stabilize learning. The environment provides real pedestrian-vehicle (AVs or HDVs) interaction trajectories as inputs. At each time step $t$, the current state $S(t)$ is constructed using spatial and kinematic features. The actor outputs a continuous action $A(t)$, which represents the pedestrian's accelerations along longitudinal and lateral directions. Each transition tuple $(S(t), A(t), R(t), S(t+\Delta t))$ is stored in a replay buffer to support off-policy learning.The model distinguishes between two types of states: $S(t)$, generated during online environment interaction, and $S(i)$, sampled from the replay buffer for training. This setup enables stable and efficient training using previously collected transitions while simulating real-time interaction behavior during inference.

### 4.2. Policy learning design

The pedestrian-centered state vector at each time step is defined as:

$$S(t) = \left(\Delta D_{\mathrm{lon}}(t), V_{\mathrm{lon}}^{\mathrm{ped}}(t), \Delta V_{\mathrm{lon}}(t), \Delta D_{\mathrm{lat}}(t), V_{\mathrm{lat}}^{\mathrm{ped}}(t), \Delta V_{\mathrm{lat}}(t)\right) \quad (9)$$

Where:

$$\Delta D_{\mathrm{lon}}(t) = Y_{\mathrm{ped}}(t) - Y_{\mathrm{veh}}(t) \quad (10)$$

$$\Delta D_{\mathrm{lat}}(t) = X_{\mathrm{ped}}(t) - X_{\mathrm{veh}}(t) \quad (11)$$

$$\Delta V_{\mathrm{lon}}(t) = V_{\mathrm{lon}}^{\mathrm{ped}}(t) - V_{\mathrm{lon}}^{\mathrm{veh}}(t) \quad (12)$$

$$\Delta V_{\mathrm{lat}}(t) = V_{\mathrm{lat}}^{\mathrm{ped}}(t) - V_{\mathrm{lat}}^{\mathrm{veh}}(t) \quad (13)$$

The action vector $A(t)$ is given by:

$$A(t) = \left(a_{\mathrm{lon}}^{\mathrm{ped}}(t), a_{\mathrm{lat}}^{\mathrm{ped}}(t)\right) \quad (14)$$

The pedestrian's action vector $A(t) = \left(a_{\mathrm{lon}}^{\mathrm{ped}}(t), a_{\mathrm{lat}}^{\mathrm{ped}}(t)\right)$ represents its accelerations along the longitudinal and lateral directions. These accelerations are used to generate the next state $S(t + \mathit{\Delta} t)$ using a deterministic kinematic update. The pedestrian's motion is updated with a fixed time interval $\Delta t = 0.1$ seconds to simulate interactions in a dynamic environment. The longitudinal and lateral velocities evolve as follows:

$$V_{\mathrm{lon}}^{\mathrm{ped}}(t+\Delta t) = V_{\mathrm{lon}}^{\mathrm{ped}}(t) + a_{\mathrm{lon}}^{\mathrm{ped}}(t) \cdot \Delta t \quad (15)$$

$$V_{\mathrm{lat}}^{\mathrm{ped}}(t+\Delta t) = V_{\mathrm{lat}}^{\mathrm{ped}}(t) + a_{\mathrm{lat}}^{\mathrm{ped}}(t) \cdot \Delta t \quad (16)$$

Relative velocities with respect to the vehicle are then updated as:

$$\Delta V_{\mathrm{lon}}(t+\Delta t) = V_{\mathrm{lon}}^{\mathrm{ped}}(t+\Delta t) - V_{\mathrm{lon}}^{\mathrm{veh}}(t) \quad (17)$$

$$\Delta V_{\mathrm{lat}}(t+\Delta t) = V_{\mathrm{lat}}^{\mathrm{ped}}(t+\Delta t) - V_{\mathrm{lat}}^{\mathrm{veh}}(t) \quad (18)$$

Next, the relative positions are updated by integrating the velocities:

$$\Delta D_{\text{lon}}(t+\Delta t) = \Delta D_{\text{lon}}(t) + \frac{V_{\text{lon}}^{\text{ped}}(t) + V_{\text{lon}}^{\text{ped}}(t+\Delta t)}{2} \cdot \Delta t \tag{19}$$

$$\Delta D_{\text{lat}}(t+\Delta t) = \Delta D_{\text{lat}}(t) + \frac{V_{\text{lat}}^{\text{ped}}(t) + V_{\text{lat}}^{\text{ped}}(t+\Delta t)}{2} \cdot \Delta t \tag{20}$$

In this formulation, the vehicle is treated as a background object with velocity derived from recorded trajectories. The pedestrian, as the learning agent, evolves solely based on its own actions. This setup enables the model to focus on learning pedestrian-initiated avoidance behavior in response to pre-defined vehicle movement patterns.

The updated state is then passed into the SMamba-based actor network, which deterministically outputs the next action $A(t) = \mu(S(t) \mid \theta^{\mu})$. The full transition tuple $(S(t), A(t), R(t), S(t+\Delta t))$ is stored in the replay buffer, which separates experience collection from parameter optimization.

During training, mini-batches of transitions $(S(i), A(i), R(i), S(i+\Delta t))$ are sampled from the buffer. The actor network generates predicted actions:

$$A(i) = \mu(S(i) \mid \theta^{\mu}) \tag{21}$$

The critic network estimates the value of the state-action pair $Q(S(i), A(i) \mid \theta^{Q})$

To stabilize learning, target actor and critic networks estimate the target values:

$$A'(i+\Delta t) = \mu'\left(S(i+\Delta t) \mid \theta^{\mu'}\right) \tag{22}$$

The parameters of the target networks are softly updated to slowly track the main networks:

$$\theta^{\mu'} \leftarrow \tau\theta^{\mu} + (1-\tau)\theta^{\mu'} \tag{23}$$

$$\theta^{Q'} \leftarrow \tau\theta^{Q} + (1-\tau)\theta^{Q'} \tag{24}$$

To guide the pedestrian agent in learning realistic and safe avoidance behaviors, this study adopts a two-level reward framework that combines step-level interaction feedback with episode-level task objectives. At the step level, each variant defines one core interaction term—either relative distance, relative velocity, or absolute velocity—to represent a specific aspect of pedestrian–vehicle dynamics. These terms encourage the pedestrian to maintain safe spacing, adjust its motion relative to the vehicle, or walk within natural speed ranges.

Distance-based reward:

$$R_D = -\left(\frac{\left|\Delta \hat{D}_{\text{lon}}(t+\Delta t) - \Delta D_{\text{lon}}(t+\Delta t)\right|}{\left|\Delta D_{\text{lon}}(t+\Delta t)\right| + \epsilon}\right) - \left(\frac{\left|\Delta \hat{D}_{\text{lat}}(t+\Delta t) - \Delta D_{\text{lat}}(t+\Delta t)\right|}{\left|\Delta D_{\text{lat}}(t+\Delta t)\right| + \epsilon}\right) \tag{25}$$

Absolute-velocity-based reward:

$$R_V = -\left(\frac{\left|\hat{V}_{\text{lon}}^{\text{ped}}(t+\Delta t) - V_{\text{lon}}^{\text{ped}}(t+\Delta t)\right|}{\left|V_{\text{lon}}^{\text{ped}}(t+\Delta t)\right| + \epsilon}\right) - \left(\frac{\left|\hat{V}_{\text{lat}}^{\text{ped}}(t+\Delta t) - V_{\text{lat}}^{\text{ped}}(t+\Delta t)\right|}{\left|V_{\text{lat}}^{\text{ped}}(t+\Delta t)\right| + \epsilon}\right) \tag{26}$$

Relative-velocity-based reward:

$$R_{\Delta V} = -\left(\frac{\left|\Delta\hat{V}_{\text{lon}}(t+\Delta t) - \Delta V_{\text{lon}}(t+\Delta t)\right|}{\left|\Delta V_{\text{lon}}(t+\Delta t)\right| + \epsilon}\right) - \left(\frac{\left|\Delta\hat{V}_{\text{lat}}(t+\Delta t) - \Delta V_{\text{lat}}(t+\Delta t)\right|}{\left|\Delta V_{\text{lat}}(t+\Delta t)\right| + \epsilon}\right) \tag{27}$$

A small constant $\epsilon = 1 \times 10^{-7}$ is added to prevent numerical instability. The notation $\Delta\hat{D}_{\text{lon}}$, $\Delta\hat{D}_{\text{lat}}$, $\hat{V}_{\text{lon}}^{\text{ped}}$, $\hat{V}_{\text{lat}}^{\text{ped}}$, $\Delta\hat{V}_{\text{lon}}$, and $\Delta\hat{V}_{\text{lat}}$ represents the predicted values from the actor network, while the unmarked variables denote the corresponding ground-truth values from recorded trajectories. Each reward term is defined as a negative normalized deviation so that smaller prediction errors yield higher rewards, encouraging accurate and stable avoidance behavior. These three formulations ($R_D, R_{\Delta V}$, and $R_V$) are trained separately to determine which reward structure best captures human-like avoidance behavior. At the episode level, two event-based rewards apply to all three variants. These rewards ensure global safety and task completion.

Collision Penalty ($R_{\text{collision}}$): A large negative reward is given immediately when a collision occurs, enforcing safety as the primary constraint:

$$R_{\text{collision}} = \begin{cases} C_{\text{collision}}, & \text{if a collision occurs} \\ 0, & \text{otherwise} \end{cases} \tag{28}$$

where $C_{\text{collision}}$ is a constant (-200) to strongly discourage unsafe maneuvers (Pérez-Gil et al. 2022).

Goal Achievement Reward ($R_{\text{goal}}$): A positive reward is assigned when the agent reaches the target location without violations:

$$R_{\text{goal}} = \begin{cases} C_{\text{goal}}, & \text{if the goal is reached} \\ 0, & \text{otherwise} \end{cases} \tag{29}$$

where $C_{\text{goal}}$ is set to 100, promoting efficient navigation (Pérez-Gil et al. 2022).

Thus, the total reward for each configuration is expressed as:

$$R_t = \begin{cases} R_{\text{collision}} & \text{if a collision occurs} \\ R_{\text{goal}} & \text{if the destination is reached} \\ R_{\text{core}} & \text{others} \end{cases} \tag{30}$$

where $R_{\text{core}}$ represents one of $\{R_D, R_{\Delta V}, R_V\}$.

This structure enables direct comparison of three SMamba-DDPG variants ($\text{SMamba} - \text{DDPG}_D$, Smamba-$\text{DDPG}_{\Delta V}$, and $\text{SMamba} - \text{DDPG}_V$) to determine which interaction feature most effectively guides realistic and safe pedestrian behavior in safety-critical situations.

The SMamba-DDPG algorithm is presented in Algorithm 1

**Algorithm 1**: SMamba-DDPG for Vehicle (AVs or HDVs)-Pedestrian Interaction Modeling (based on (Lillicrap *et al.* 2015, Gu and Dao 2023, Xu *et al.* 2025))

**Input**: Driving dataset containing Vehicle-Pedestrian interaction trajectories
**Output**: Learned pedestrian policy $\mu(S(t))$ for collision avoidance

Initialize actor network $\mu(S(t) \mid \theta^{\mu})$ and critic network $Q(S(t), A(t) \mid \theta^{Q})$ with random weights $\theta^{\mu}$, $\theta^{Q}$

Initialize target network $\theta^{Q'} \leftarrow \theta^{Q}, \theta^{\mu'} \leftarrow \theta^{\mu}$.

Initialize replay buffer $\mathcal{D}$

For each episode $e = 1,2, \ldots, M$:

Observe initial state $S(1) = \left[\Delta D_{\text{ped}}^{\text{lon}}(1), V_{\text{ped}}^{\text{lon}}(1), \Delta V_{\text{ped}}^{\text{lon}}(1), \Delta D_{\text{ped}}^{\text{lat}}(1), V_{\text{ped}}^{\text{lat}}(1), \Delta V_{\text{ped}}^{\text{lat}}(1)\right]$

Initialize exploration noise process $\mathcal{N}$

For each time step $t = 1,2, \ldots, T$ :

Pass $S(t)$ through SMamba actor network, including selective state modeling and SMOOTH modules

Select action with exploration:

$$A(t) = \mu(S(t) \mid \theta^{\mu}) + \mathcal{N}(t), \text{ where } A(t) = \left[a_{\text{ped}}^{\text{lon}}(t), a_{\text{ped}}^{\text{lat}}(t)\right]$$

Update pedestrian state using the kinematic model

$$S(t + \Delta t) \leftarrow \text{Simulate } (S(t), A(t))$$

Compute reward $R(t)$ using combined reward function (distance, velocity, and velocity difference)

Store transition $(S(t), A(t), R(t), S(t + \Delta t))$ in buffer $\mathcal{D}$

Sample mini-batch $\{(S(i), A(i), R(i), S(i + \Delta t))\}_{i=1}^{N} \sim \mathcal{D}$

Compute target value:

$$y_i = R(i) + \gamma \cdot Q'\left(S(i + \Delta t), \mu'\left(S(i + \Delta t) \mid \theta^{\mu'}\right) \mid \theta^{Q'}\right)$$

Update critic by minimizing loss:

$$\mathcal{L} = \frac{1}{N} \sum_{i} \left(y_i - Q(S(i), A(i) \mid \theta^{Q})\right)^2$$

Update actor via policy gradient:

$$\nabla_{\theta^{\mu}} J \approx \frac{1}{N} \sum_{i} \nabla_a Q(S, A \mid \theta^{Q}) \Big|_{S=S(i), A=\mu(S(i))} \cdot \nabla_{\theta^{\mu}} \mu(S \mid \theta^{\mu}) \Big|_{S=S(i)}$$

Update target networks using soft update rule:

$$\theta^{Q'} \leftarrow \tau\theta^{Q} + (1-\tau)\theta^{Q'}$$
$$\theta^{\mu'} \leftarrow \tau\theta^{\mu} + (1-\tau)\theta^{\mu'}$$

End for (time steps)
End for (episodes)

## 5. Results and Discussion

### 5.1. SMamba-DDPG framework training and results

The SMamba-DDPG framework was trained on safety-critical pedestrian–vehicle interaction data identified using the CurvTTC criterion. For Pedestrian-AV scenarios, 411 safety-critical cases (8,865 time steps) were extracted; for Pedestrian-HDV interactions, 2,404 cases (52,895 time steps) were identified. Each dataset was split into 80% for training and 20% for testing. Two separate sets of model weights were trained to capture distinct behavioral responses. One set models pedestrian responses to AVs, and the other models responses to HDVs.

Figure 6 presents the architecture of the actor and critic networks used in the SMamba-DDPG framework. The model was trained separately on Pedestrian-AV and Pedestrian-HDV datasets, producing two distinct sets of parameters. This design allows the pedestrian agent to learn distinct behavioral responses to AV and HDV. The pedestrian agent receives the six-dimensional state vector $S(t)$ defined in Eq. (9), which encodes the relative and absolute kinematic features required for sequential decision-making. In the actor network, the state is first processed through a Selective Mamba (SMamba) module, which extracts temporal features using structured state space modeling. These features are passed through two fully connected layers of 256 neurons each. The output layer generates a continuous two-dimensional action:

$$A(t) = \left(a_{\text{lon}}^{\text{ped}}(t), a_{\text{lat}}^{\text{ped}}(t)\right) \tag{31}$$

which represents the predicted accelerations along the longitudinal and lateral axes. The critic network receives both the state $S(t)$ and the action $A(t)$ as inputs. The state branch processes the input through a SMamba block, followed by two fully connected layers with 16 and 32 neurons. In parallel, the action branch processes $A(t)$ using two layers with the same configuration. The outputs from both branches are concatenated and fed into two fully connected layers with 256 neurons. Finally, the critic outputs a scalar $Q(S(t), A(t) \mid \theta^{Q})$, estimating the long term return for taking action $A(t)$ under state $S(t)$.

This design allows the agent to focus on risk-relevant motion features and produce smooth, realistic avoidance behaviors. By training the actor-critic networks separately for AV and HDV scenarios, the framework captures the nuanced differences in pedestrian decision-making when interacting with automated versus human-controlled vehicles.

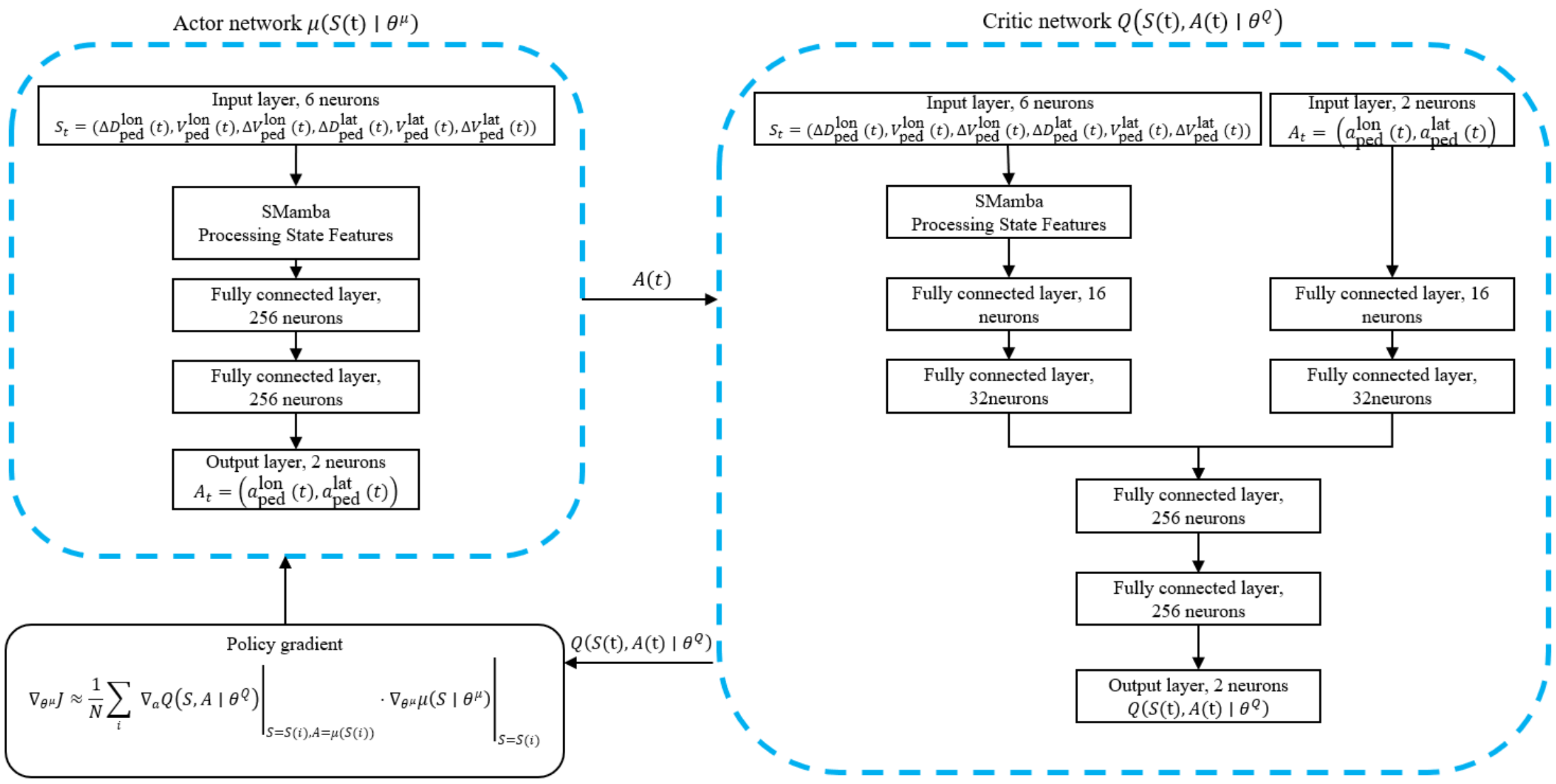

**Figure 6: Actor and critic networks of SMamba-DDPG**

The SMamba-DDPG framework uses a 1-second sliding memory window to capture sequential decision-making patterns. A trajectory masking mechanism was employed to manage incomplete early-stage sequences, enabling the model to focus on high-risk, behaviorally informative time steps. The actor and critic networks were trained over 3,000 episodes, with network stability ensured by applying soft target updates at regular intervals. The training settings are summarized in Table 3.

**Table 3 Hyperparameters Used for Training the SMamba-DDPG Framework**

| Hyperparameter | Value | Description |
|---|---|---|
| Buffer capacity | 10,000 | Size of replay buffer for storing interaction sequences |
| Batch size | 256 | Number of samples per training iteration |
| Learning rate (Actor) | 0.0005 | Step size for updating actor network |
| Learning rate (Critic) | 0.001 | Step size for updating critic network |
| Discount factor $\gamma$ | 0.9 | Discount rate for future rewards |
| Target update rate $\tau$ | 0.01 | Exponential moving average rate for target network updates |
| Exploration noise $\sigma$ | 0.01 | Gaussian noise added to action during exploration |
| Total episodes | 3,000 | Total number of training episodes |
| Reward scaling (Actor) | 100 | Scaling factor applied to actor's reward signal |
| Reward scaling (Critic) | 100 | Scaling factor applied to critic's reward signal |

Three SMamba-DDPG models were developed using distinct reward components: distance-based $Reward_D$), relative velocity-based ($Reward_{\Delta V}$), and absolute velocity-based ($Reward_v$) to investigate the influence of different reward structures on training performance. These models capture different aspects of pedestrian decision-making during vehicle interactions. Separate versions were trained for Pedestrian-AV and Pedestrian-HDV scenarios to reflect potential behavioral differences. Figure 7 presents the rolling average episodic reward curves over 3,000 episodes (window size = 50).

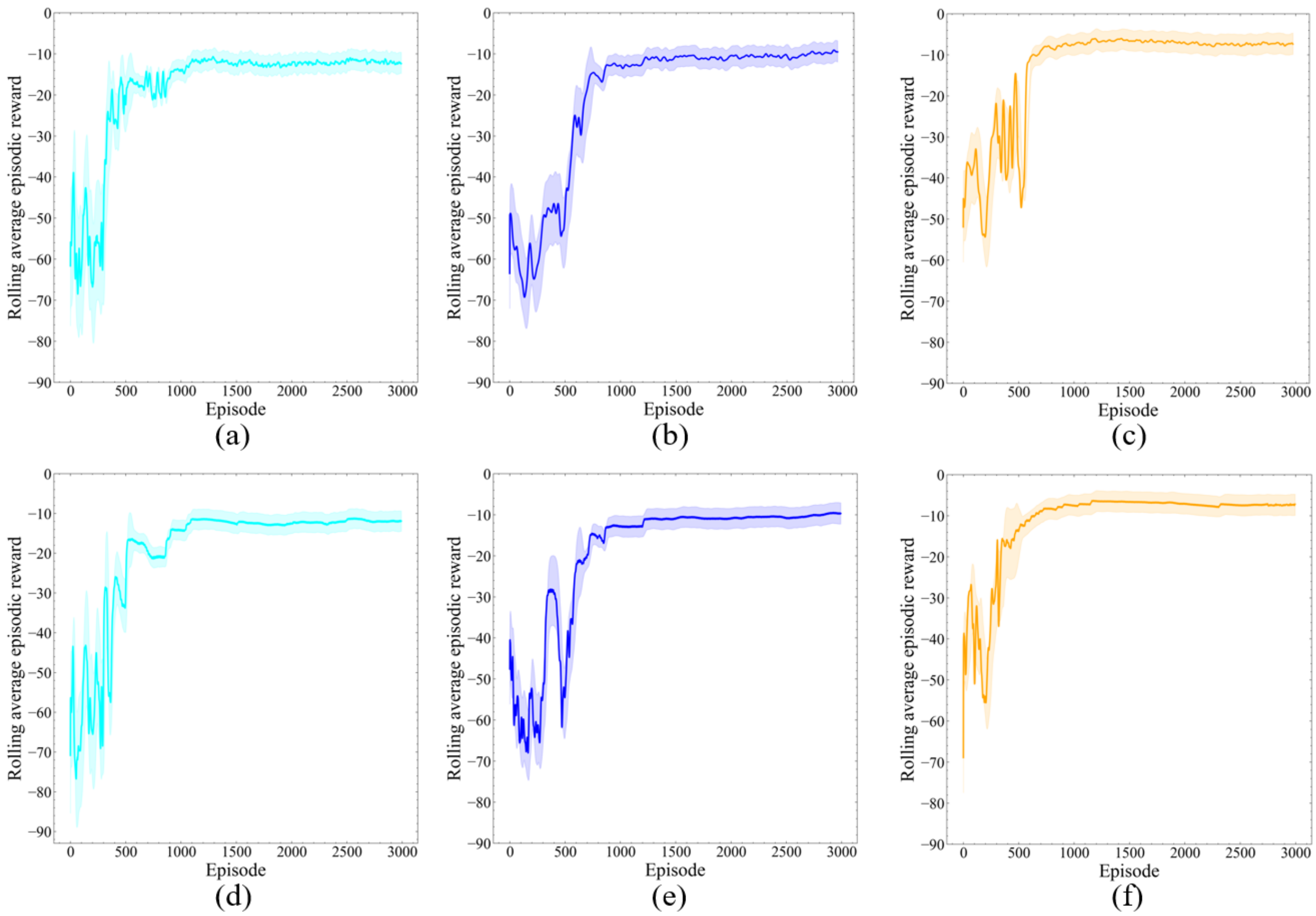

**Figure 7**: **Training performance of SMamba-DDPG variants for AV and HDV models. (a–c) AV-based models; (d–f) HDV-based models. Each column represents variants with different state inputs: (a, d) distance-based ($\mathrm{SMamba-DDPG}_D$), (b, e) relative-velocity-based ($\mathrm{SMamba-DDPG}_{\Delta V}$), and (c, f) velocity-based ($\mathrm{SMamba-DDPG}_v$)**

Among all models, $\mathrm{SMamba-DDPG}_v$ exhibited the highest and most stable reward trajectory. This model converged quickly with reduced variance after 500 episodes, indicating consistent learning. In contrast, $\mathrm{SMamba-DDPG}_D$ showed slower convergence and larger fluctuations. $\mathrm{SMamba-DDPG}_{\Delta V}$ achieved moderate stability but lower peak rewards compared to the velocity-based model.

These results demonstrate that the absolute velocity reward provides the most effective learning signal, enabling the pedestrian agent to learn safer and more stable collision avoidance behaviors. The shaded regions represent confidence intervals across multiple training runs, highlighting the robustness of each reward formulation.

This study computed the root mean square error (RMSE) (Willmott and Matsuura 2005) across key motion variables to evaluate how accurately each model reproduces pedestrian responses under AV and HDV interactions. The RMSE measures the average deviation between predicted and observed values, defined as:

$$RMSE = \sqrt{\frac{1}{n}\sum_{i=1}^{n} (\hat{x}_i - x_i)^2} \tag{32}$$

where $\hat{x}_i$ and $x_i$ are the predicted and actual values at time step $i$, respectively. Table 4 reports RMSE values for six variables: pedestrian velocity, acceleration, and relative distance, each in the longitudinal and lateral directions. These values were computed separately for Pedestrian-AV and Pedestrian-HDV cases.

Three variants of the SMamba-DDPG model were evaluated using different reward signals: distance-based ($\mathrm{SMamba-DDPG}_D$), velocity-based ($\mathrm{SMamba-DDPG}_v$), and relative-velocity-based ($\mathrm{SMamba-DDPG}_{\Delta V}$). Among these, the velocity-based model ($\mathrm{SMamba-DDPG}_v$) consistently achieved the lowest RMSE across most variables, for both AV and HDV scenarios. Notably, it produced the smallest errors in longitudinal and lateral velocity, acceleration, and distance, indicating improved fidelity in modeling pedestrian motion.

These results demonstrate that the velocity-based reward signal provides more effective behavioral guidance, helping the agent learn smoother and more realistic acceleration profiles. In contrast, the models trained with distance-based and relative-velocity-based rewards showed higher errors, suggesting weaker alignment with natural pedestrian behavior.

**Table 4 RMSE between model predictions and real data under different reward conditions**

| Variable | $\mathrm{SMamba-DDPG}_D$ | | $\mathrm{SMamba-DDPG}_v$ | | $\mathrm{SMamba-DDPG}_{\Delta V}$ | |
|---|---|---|---|---|---|---|
| | Pedestrian-AV | Pedestrian-HDV | Pedestrian-AV | Pedestrian-HDV | Pedestrian-AV | Pedestrian-HDV |
| $V_{\mathrm{ped}}^{\mathrm{lon}}$ | 0.71 | 0.69 | 0.62* | 0.63* | 0.72 | 0.68 |
| $a_{\mathrm{ped}}^{\mathrm{lon}}$ | 0.82 | 0.77 | 0.59* | 0.55* | 0.75 | 0.73 |
| $\Delta D_{\mathrm{ped}}^{\mathrm{lon}}$ | 1.08 | 1.19 | 0.86* | 0.84* | 1.05 | 1.14 |
| $V_{\mathrm{ped}}^{\mathrm{lat}}$ | 0.73 | 0.75 | 0.45* | 0.47* | 0.69 | 0.72 |
| $a_{\mathrm{ped}}^{\mathrm{lat}}$ | 0.68 | 0.72 | 0.67* | 0.68* | 0.83 | 0.79 |
| $\Delta D_{\mathrm{ped}}^{\mathrm{lat}}$ | 0.82 | 0.91 | 0.69* | 0.70* | 0.87 | 0.83 |

* indicates the best value for each metric.

The SMamba-DDPG framework was evaluated against a series of baseline models to examine its effectiveness in reproducing pedestrian crash avoidance behaviors in both Pedestrian-AV and Pedestrian-HDV interactions. Model performance was compared using three key indicators: the RMSE, which measures average prediction deviation; the Average Displacement Error (ADE), which evaluates the mean spatial deviation across trajectories; and the Final Displacement Error (FDE), which assesses the positional error at the trajectory endpoint. ADE and FDE are defined as:

$$ADE = \frac{1}{nT}\sum_{i=1}^{n}\sum_{t=1}^{T} \left\|\hat{y}_{i,t} - y_{i,t}\right\|_2 \tag{33}$$

$$FDE = \frac{1}{n}\sum_{i=1}^{n} \left\|\hat{y}_{i,T} - y_{i,T}\right\|_2 \tag{34}$$

where $n$ is the number of trajectories, $T$ is the prediction horizon, and $\hat{y}_{i,t}$ and $y_{i,t}$ are the predicted and ground-truth positions for trajectory $i$ at time step $t$.

The comparison included supervised learning models, standard deep reinforcement learning (DRL) algorithms, and hybrid DRL variants that incorporated temporal or state-space representations. The results are summarized in Table 5.

Among all models, supervised approaches such as Neural Network (NN), LSTM, and Transformer exhibited the weakest performance. The NN baseline, implemented with three fully connected layers and ReLU activations, produced the largest errors (e.g., $RMSE_{V_{\text{ped}}^{\text{lon}}} = 3.68\ \text{m/s}$ ), indicating limited adaptability to dynamic pedestrian-vehicle interactions. LSTM showed similar weakness, while the Transformer remained limited despite slightly lower RMSEs. These supervised models were trained using 10-step state sequences in PyTorch, optimized with the Adam algorithm (learning rate $= 0.001$, batch size $= 128$ ), and regularized through early stopping based on validation loss.

Standard DRL algorithms, including DDPG, DQN, and DPG, achieved moderate improvements but still underperformed in safety-critical situations requiring rapid responses. For instance, DDPG recorded $RMSE_{V_{\text{ped}}^{\text{lon}}} = 1.57\ \text{m/s}$ and $RMSE_{a_{\text{ped}}^{\text{lon}}} = 1.74\ \text{m/s}$, suggesting difficulty in modeling abrupt and time-sensitive pedestrian reactions.

Hybrid temporal models (Transformer-DDPG, LSTM-DDPG, and Mamba-DDPG) improved accuracy by capturing temporal dependencies in policy learning. The Mamba-DDPG model benefited from the state-space formulation, which enhanced noise filtering and long-range dependency capture, resulting in a longitudinal velocity RMSE of about $0.67\ \text{m/s}$.

The SMamba-DDPG achieved the best overall performance across all kinematic indicators. For both Pedestrian-AV and Pedestrian-HDV settings, it consistently yielded the lowest RMSE (e.g., $RMSE_{V_{\text{ped}}^{\text{lon}}} = 0.62$ and $0.63\ \text{m/s}$, respectively) and superior ADE/FDE values ($0.65$ m and $1.12$ m). These results confirm the framework's robustness and generalizability. The performance gain is primarily attributed to the SMamba block's smoothing mechanism, which stabilizes hidden-state transitions and enhances temporal coherence.

Table 5 Comparative analysis of RMSE across different Models

| Category | Scenarios | Model | RMSE $V_{\mathrm{ped}}^{\mathrm{lon}}$ (m/s) | RMSE $a_{ped}^{lon}$ (m/s²) | RMSE $V_{\mathrm{ped}}^{\mathrm{lat}}$ (m/s) | RMSE $a_{ped}^{lat}$ (m/s²) | ADE (m) | FDE (m) |
|---|---|---|---|---|---|---|---|---|
| Hybrid Models | Pedestrian-AV | SMamba-DDPG | 0.62* | 0.59 | 0.45* | 0.67* | 0.65 | 1.12* |
| | | Mamba-DDPG | 0.67 | 0.68 | 0.61 | 0.86 | 0.72 | 1.25 |
| | | Transformer-DDPG | 0.79 | 0.94 | 0.83 | 1.18 | 0.93 | 1.63 |
| | | LSTM-DDPG | 1.19 | 1.07 | 1.06 | 1.83 | 1.18 | 2.17 |
| | Pedestrian-HDV | SMamba-DDPG | 0.63 | 0.55* | 0.47 | 0.68 | 0.64* | 1.13 |
| | | Mamba-DDPG | 0.68 | 0.70 | 0.56 | 0.89 | 0.71 | 1.28 |
| | | Transformer-DDPG | 0.81 | 0.97 | 0.82 | 1.11 | 0.95 | 1.64 |
| | | LSTM-DDPG | 1.16 | 1.03 | 1.09 | 1.87 | 1.21 | 2.15 |
| Reinforcement Learning Baselines | Pedestrian-AV | DDPG | 1.57 | 1.74 | 1.29 | 1.52 | 1.42 | 2.44 |
| | | DQN | 1.79 | 1.99 | 1.55 | 1.88 | 1.78 | 3.08 |
| | | DPG | 2.22 | 2.11 | 2.06 | 2.64 | 2.32 | 4.12 |
| | Pedestrian-HDV | DDPG | 1.65 | 1.83 | 1.26 | 1.62 | 1.55 | 2.57 |
| | | DQN | 1.78 | 2.01 | 1.54 | 1.97 | 1.77 | 3.14 |
| | | DPG | 2.34 | 2.09 | 2.17 | 2.58 | 2.44 | 4.13 |
| Supervised Learning Baselines | Pedestrian-AV | Transformer | 1.95 | 1.96 | 2.17 | 2.44 | 2.51 | 4.52 |
| | | LSTM | 3.25 | 2.91 | 3.99 | 3.97 | 3.25 | 5.95 |
| | | NN | 3.68 | 3.01 | 3.72 | 3.28 | 3.27 | 6.41 |
| | Pedestrian-HDV | Transformer | 2.01 | 2.07 | 2.09 | 2.37 | 2.43 | 4.25 |
| | | LSTM | 3.19 | 3.08 | 3.92 | 3.88 | 3.44 | 6.08 |
| | | NN | 3.57 | 3.16 | 3.93 | 3.18 | 3.56 | 6.34 |

* The best values among all the models.

### 5.2. Reconstruction of avoidance trajectories

This section evaluates how well the $\mathrm{SMamba} - \mathrm{DDPG}_v$ framework reconstructs pedestrian avoidance trajectories when interacting with AVs and HDVs. The model generated reconstructed trajectories sequentially using each pedestrian's initial state as input, allowing the model to reproduce realistic motion dynamics under both conditions.

To quantitatively evaluate reconstruction accuracy, Table 6 summarizes the RMSE for six motion metrics in both Pedestrian-AV and Pedestrian-HDV scenarios. The low RMSE values across all variables (longitudinal velocity: 0.28 m/s for AV, 0.19 m/s for HDV; longitudinal acceleration: 0.33 m/s² for AV, 0.35 m/s² for HDV) confirm that the model accurately reproduces pedestrian kinematics in both AV and HDV scenarios, consistent with the trajectory characteristics in Figure 8.

**Table 6 RMSE of Pedestrian Kinematics Reconstruction under AV and HDV Conditions**

| Metric | Pedestrian-AV | Pedestrian-HDV |
|---|---|---|
| Longitudinal Distance RMSE (m) | 0.16 | 0.19 |
| Lateral Distance RMSE (m) | 0.25 | 0.24 |
| Longitudinal Speed RMSE (m/s) | 0.28 | 0.19 |
| Lateral Speed RMSE (m/s) | 0.34 | 0.40 |
| Longitudinal Acceleration RMSE (m/s²) | 0.33 | 0.35 |
| Lateral Acceleration RMSE (m/s²) | 0.31 | 0.29 |

These results highlight two key findings. First, $\mathrm{SMamba} - \mathrm{DDPG}_v$ can reliably reconstruct realistic pedestrian responses in both AV and HDV scenarios. Second, the pedestrian behavior diverges notably depending on the vehicle type, underscoring the necessity of using separate models or policy weights for AV and HDV conditions in safety-critical trajectory prediction tasks.

Figure 8 presents four representative safety-critical interaction scenarios involving pedestrians and vehicles. Subfigures (a) and (b) depict Pedestrian-AV interactions, while (c) and (d) correspond to Pedestrian-HDV cases. Each row illustrates a complete 11-second interaction, including four synchronized plots: the CurvTTC curve, lateral-longitudinal movement map, longitudinal position over time, and speed profile of both agents. Although the full trajectories are displayed, the $\mathrm{SMamba} - \mathrm{DDPG}_v$ framework reconstructs pedestrian behavior only during high-risk intervals, defined as periods when CurvTTC < 5 s. These intervals occur during sudden vehicle maneuvers that require rapid pedestrian responses. The reconstructed pedestrian crash avoidance behaviors (shown in red dashed lines) focus on these risk-critical intervals rather than the entire interaction span.

In Pedestrian-AV scenarios, the CurvTTC curves drop as AVs decelerate or slightly deviate from their paths. Pedestrians respond defensively by reducing speed, pausing, or shifting laterally to avoid potential conflicts. The $\mathrm{SMamba} - \mathrm{DDPG}_v$ model accurately reproduces these adaptive behaviors, closely matching both the timing and spatial pattern of observed reactions. For Pedestrian-HDV encounters, pedestrians exhibit stronger avoidance responses as HDVs maintain forward motion or execute turning maneuvers. Sudden drops in CurvTTC trigger rapid pedestrian deceleration or lateral deviations. The reconstructed trajectories align well with real behaviors, demonstrating that $\mathrm{SMamba} - \mathrm{DDPG}_v$ effectively captures human-like decision-making under high-risk, time-critical conditions.

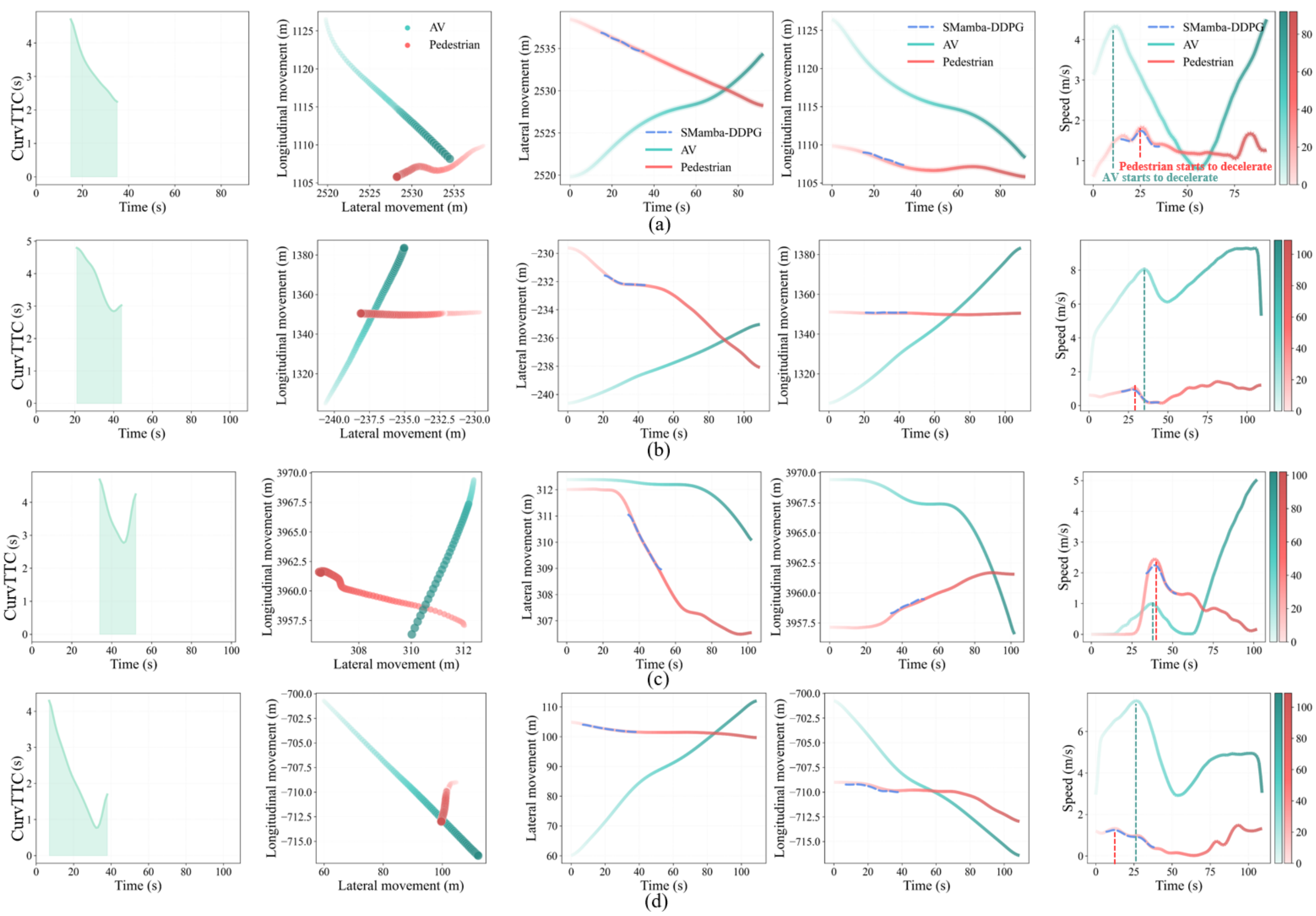

**Figure 8: Example of trajectory reconstruction: (a)-(b) Pedestrian-AV, (c)-(d) Pedestrian-HDV**

These findings confirm that vehicle behavior strongly influences pedestrian decision-making during safety-critical situations. Rather than modeling entire trajectories, $\mathrm{SMamba} - \mathrm{DDPG}_v$ focuses on safety-critical scenarios. This targeted reconstruction approach allows the model to generate responses that closely match observed kinematic patterns in real pedestrian trajectories.

## 5.3. Validate the reconstructed data by analyzing reaction times

This section validates the reconstructed data by comparing model-generated and real-world reaction times for pedestrian–vehicle safety-critical interactions. We evaluate whether the SMamba-DDPG framework reproduces realistic response delays in Argoverse-2 AV and HDV scenarios by aligning reconstructed trajectories with matched real events and analyzing their reaction-time distributions.

Real-world pedestrian–vehicle interactions were extracted from 411 safety-critical events in the Argoverse 2 dataset using the CurvTTC criterion. An equal number of pedestrian-HDV cases were randomly selected from the same dataset for balanced comparison. Two corresponding sets of reconstructed trajectories were generated using separately trained SMamba-DDPG models for AV and HDV interactions, ensuring one-to-one consistency with observed data.

Reaction time is defined as the delay between a leading agent's maneuver and a following agent's corresponding response (Ma and Andréasson 2006, Mehmood and Easa 2009). For pedestrian–vehicle interactions, this delay is computed using both longitudinal and lateral dimensions:

$$\text{Observed delay } = \arg\min_{t_d} \left\| \begin{bmatrix} a_{\text{lon}}^{\text{ped}}(t) - a_{\text{lon}}^{\text{veh}}(t - t_d) \\ a_{\text{lat}}^{\text{ped}}(t) - a_{\text{lat}}^{\text{veh}}(t - t_d) \end{bmatrix} \right\|_2 \text{ subject to } \left\| \begin{bmatrix} a_{\text{lon}}^{\text{ped}}(t) - a_{\text{lon}}^{\text{ped}}(t - 1) \\ a_{\text{lat}}^{\text{ped}}(t) - a_{\text{lat}}^{\text{ped}}(t - 1) \end{bmatrix} \right\|_2 > \varepsilon \quad (35)$$

where

- $a_{\text{lon}}^{\text{veh}}\,(t - t_d)$ and $a_{\text{lat}}^{\text{veh}}\,(t - t_d)$ are the vehicle's longitudinal and lateral accelerations at time $t - t_d$,
- $\|\cdot\|_2$ denotes the Euclidean (L2) norm,
- $\varepsilon = 0.05\ \text{m/s}^2$ (Ozaki, 1993) is the threshold to exclude minor fluctuations,
- and $t_d$ is the observed reaction delay in frames (with 10 Hz sampling rate, one frame $= 0.1$ s).

Using this method, 411 reaction times were computed for each interaction set. Figure 9 illustrates their distributions for both real-world and reconstructed safety-critical interactions.

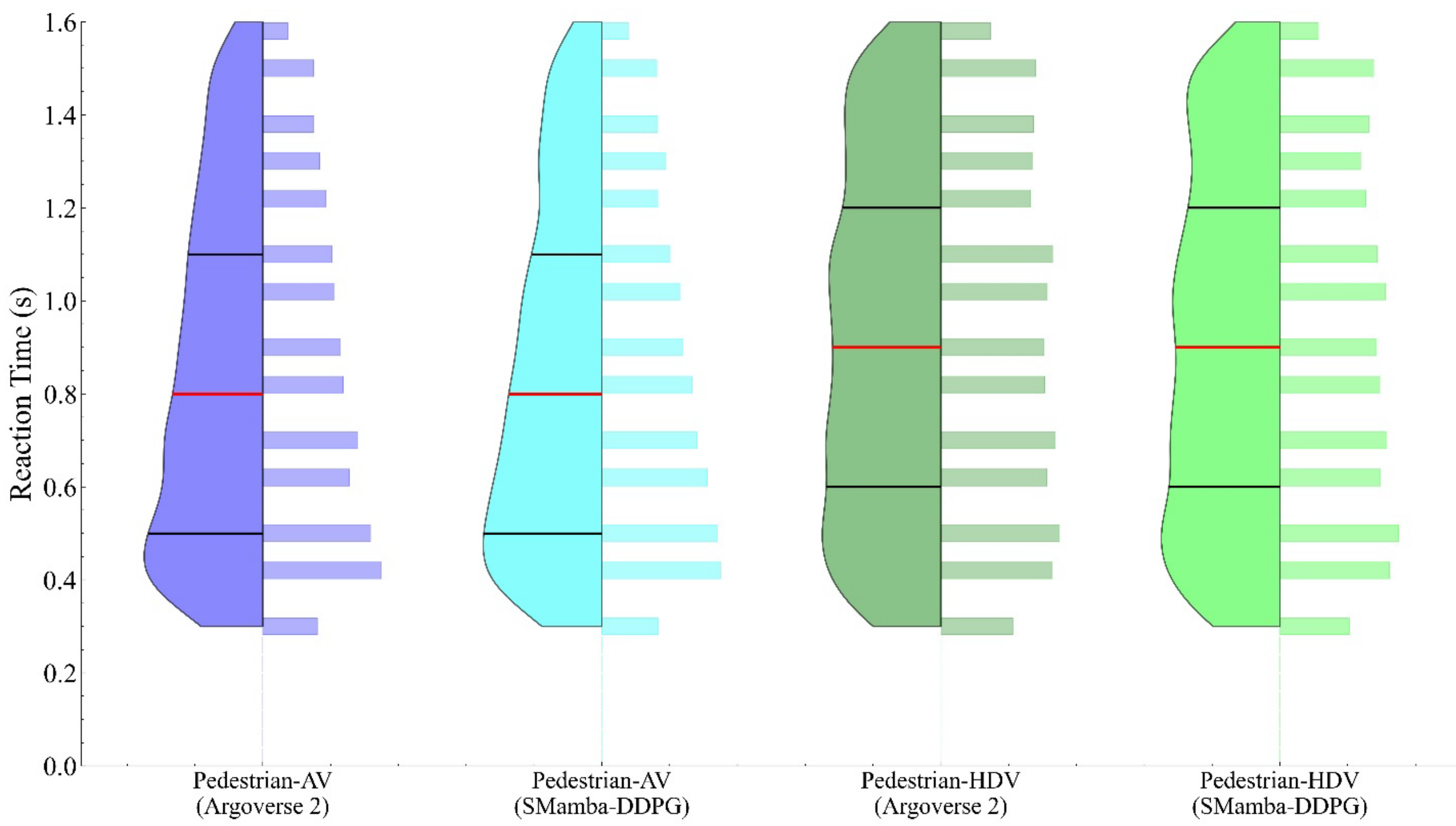

**Figure 9: Reaction time distributions of pedestrian–AV and pedestrian–HDV interactions for Argoverse 2 and SMamba-DDPG reconstructed data**

Figure 9 presents reaction time distributions using panels that each combine a half-violin plot showing the KDE density shape, a histogram displaying frequency counts, and embedded quartile lines (red: mean; black: first and third quartiles). For Pedestrian-AV (Argoverse 2), real interaction times are concentrated between 0.4 and 0.8 seconds, with the KDE curve showing a steep drop at higher values, indicating relatively quick and consistent pedestrian responses. The Pedestrian-AV (SMamba-DDPG) panel closely replicates this pattern, with mean and interquartile range aligning well with the observed data, though with a slightly broader spread. For pedestrian–HDV interactions, both Pedestrian-HDV (Argoverse 2) and Pedestrian-HDV (SMamba-DDPG) display wider and more irregular distributions, reflecting the greater variability in pedestrian responses to human-driven vehicles. Pedestrian reactions to AVs show lower reaction times compared to HDVs. Across all four panels, the mean and interquartile ranges of reconstructed data closely match those of real interactions, confirming that SMamba-DDPG reproduces realistic response timing for both AV and HDV scenarios.

To statistically evaluate differences in reaction times, a two-sample t-test was used to compare mean values, and a Kolmogorov–Smirnov (K–S) test assessed distribution similarity. The results are summarized in Table 7.

**Table 7 Statistical Comparison of Reaction Times Across Interaction Types**

| Group 1 | Group 2 | t-statistic | p-value | KS statistic | p-value | Significant Difference |
|---|---|---|---|---|---|---|
| (i)Pedestrian-AV (Argoverse 2) | (ii)Pedestrian-AV (SMamba-DDPG) | 0.1060 | 0.9156 | 0.0089 | 0.9897 | No |
| (iii) Pedestrian-HDV (Argoverse 2) | (iv) Pedestrian-HDV (SMamba-DDPG) | 1.2058 | 0.2279 | 0.0128 | 0.8007 | No |
| (i)Pedestrian-AV (Argoverse 2) | (iii) Pedestrian-HDV (Argoverse 2) | -9.0527 | <0.001 | 0.0813 | <0.001 | Yes |
| (ii)Pedestrian-AV (SMamba-DDPG) | (iv) Pedestrian-HDV (SMamba-DDPG) | -7.9516 | <0.001 | 0.0782 | <0.001 | Yes |

The reconstructed SMamba-DDPG results closely align with the real-world Argoverse-2 data, showing no significant statistical differences. Pedestrian reactions to AVs show lower reaction times compared to HDVs, which display wider and more irregular distributions, confirming that the model reproduces human-like response delays under safety-critical conditions. This shorter response latency does not necessarily imply reduced caution. A possible model-based interpretation is that pedestrian kinematics respond earlier to AV motion cues, while subsequent crossing speeds remain comparatively lower.

## 5.4. Counterfactual scenario analysis: quantifying pedestrian behavioral differences under AVs vs HDVs

### 5.4.1 Pedestrian–AV model applied to HDV scenarios

This section examines how the SMamba-DDPG pedestrian model—trained on pedestrian–AV interactions—responds when exposed to HDV trajectories. Two scenarios are compared in this analysis:

- **Observation (Pedestrian-HDV)**: Baseline pedestrian behavior in real-world safety-critical events with HDVs, extracted from the Argoverse 2 dataset.
- **Counterfactual**: The pedestrian model trained on AV interactions is evaluated within HDV trajectories from the Argoverse 2 dataset. The model produces pedestrian responses as if interacting with HDVs, using HDV motion data to compute relative distance, speed, and acceleration features.

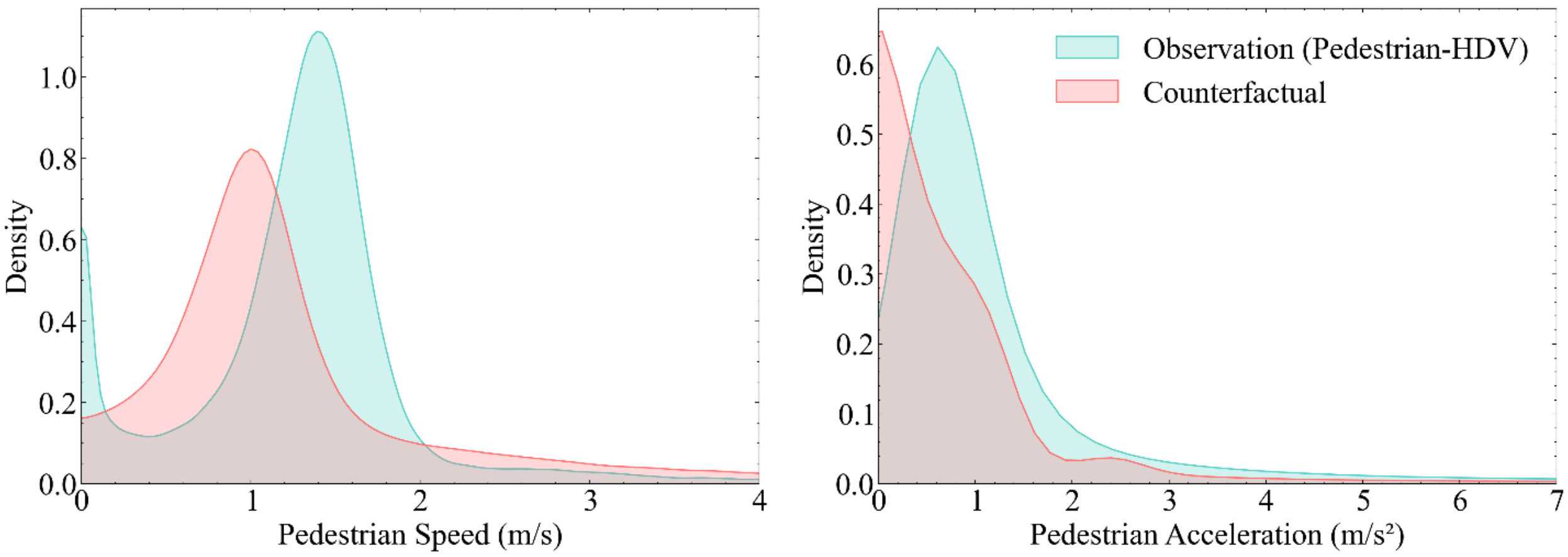

**Figure 10: Distributions of observation (Pedestrian-HDV) and counterfactual: (a) speed (b) acceleration**

Figure 10 (a) shows that the counterfactual case (AV-trained model applied to HDV) exhibits lower pedestrian speeds, concentrated between 0.5–1.5 m/s, while the observation peaks between 1–2 m/s. This suggests that the AV-trained pedestrian policy produces lower speeds even when applied to HDV trajectories. Figure 10 (b) shows that ground-truth HDV interactions cluster near 0–1 m/s². This pattern implies that pedestrians trained on AV interactions exhibit smaller acceleration fluctuations when encountering HDVs.

Table 8 provides statistical confirmation of the differences observed in Figure 10. For both speed and acceleration, the K–S tests show significant discrepancies ($p < 0.001$). Compared to the observation, the counterfactual case shows a 14.7% reduction in mean speed and a 25.7% increase in variability. This indicates a more conservative response pattern generated by the transferred policy under the evaluated simulation settings. For acceleration, the observation shows a higher mean value (2.91 m/s²) and greater variability (Std Dev = 5.82 m/s²) than the counterfactual (mean = 1.44 m/s², Std Dev = 4.20 m/s²). This reduction indicates that pedestrians trained under AV interactions respond more smoothly and with fewer abrupt speed changes when exposed to HDV trajectories.

**Table 8: Statistical Comparison of Key Driving Metrics in Observation and Counterfactual Scenarios**

| Metric | Scenario | K-S statistic | p-value | Mean | Std Dev |
|---|---|---|---|---|---|
| Pedestrians Speed (m/s) | Observation | 0.29 | < 0.001 | 1.28 | 0.72 |
| | Counterfactual | | | 1.20 | 1.17 |
| Pedestrians Acceleration (m/s²) | Observation | 0.52 | < 0.001 | 2.91 | 5.82 |
| | Counterfactual | | | 1.44 | 4.20 |

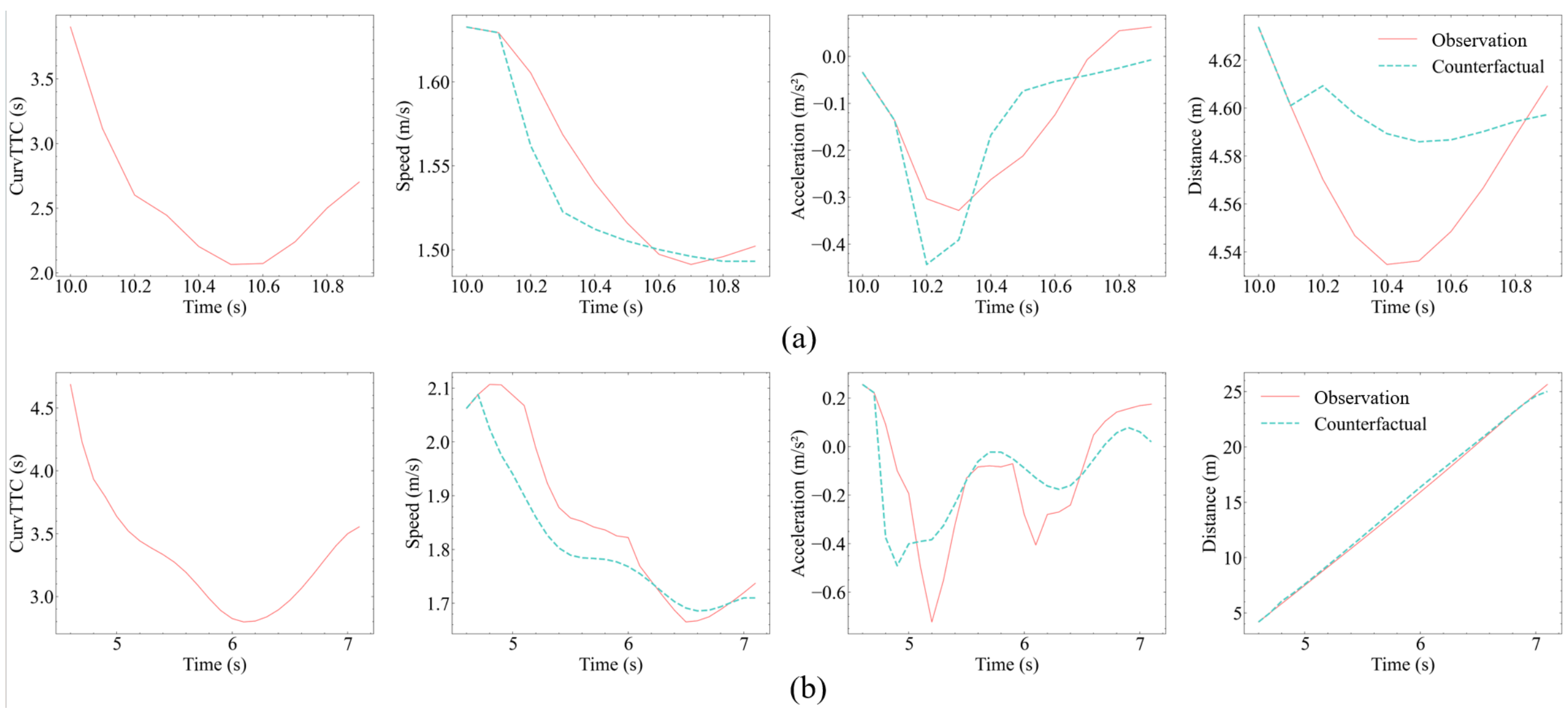

**Figure 11: Case studies of counterfactual pedestrians trajectory reconstruction**

Figure 11 illustrates representative case studies of counterfactual trajectory reconstruction. The counterfactual trajectories feature lower speeds and a higher frequency of deceleration compared with the ground-truth trajectories. In case (b), the counterfactual trajectory shows clear speed drops around 4.8 seconds, which do not appear in the Observation.

5.4.2 Pedestrian–HDV model applied to AV scenarios

This section examines how the pedestrian model trained on HDV interactions responds when applied to AV trajectories. Two scenarios are compared in this analysis:

- **Observation (Pedestrian-AV)**: Baseline pedestrian behavior in real-world safety-critical events with AVs, extracted from the Argoverse 2 dataset.
- **Counterfactual**: The pedestrian model trained on HDV interactions is evaluated within AV trajectories from the Argoverse 2 dataset.

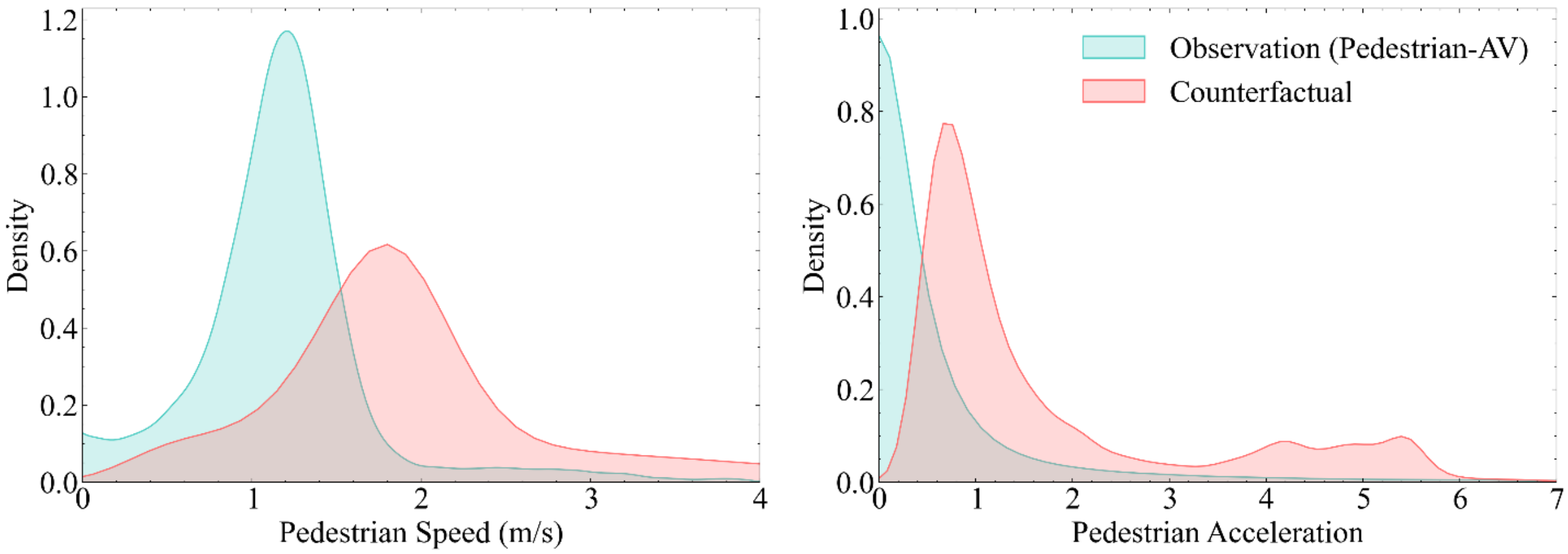

**Figure 12: Distributions of observation (Pedestrian-AV) and counterfactual: (a) speed (b) acceleration**

Figure 12 (a) shows that the counterfactual case (HDV-trained model applied to AV) results in higher pedestrian speeds, concentrated between 1.5–2.5 m/s, compared with the observation peak between 1.0–1.5 m/s. Figure 12 (b) shows that acceleration in ground-truth AV interactions is concentrated near 0–0.5 m/s², while the counterfactual displays larger positive values (0.6–0.8 m/s²) with a longer upper tail up to 6–7 m/s². Overall, the counterfactual scenario shows higher speeds and greater acceleration variability than the observation.

Table 9 confirms these trends, with significant K–S differences ($p < 0.001$) for both speed and acceleration. The counterfactual model shows higher mean pedestrian speeds (2.17 vs. 1.08 m/s) but lower mean acceleration (1.12 vs. 1.38 m/s²), suggesting that the HDV-trained policy produces faster crossing speeds yet with smoother kinematics when exposed to AV trajectories.

**Table 9: Statistical Comparison of Key Driving Metrics in Observation and Counterfactual Scenarios**

| Metric | Scenario | K-S statistic | p-value | Mean | Std Dev |
|---|---|---|---|---|---|
| Pedestrians Speed (m/s) | Observation | 0.21 | < 0.001 | 1.08 | 0.62 |
| | Counterfactual | | | 2.17 | 3.05 |
| Pedestrians Acceleration (m/s²) | Observation | 0.52 | < 0.001 | 1.38 | 2.61 |
| | Counterfactual | | | 1.12 | 2.01 |

(a)

(b)

**Figure 13: Case studies of counterfactual pedestrian trajectory reconstruction**

Figure 13 presents representative counterfactual cases demonstrating that pedestrians exhibit faster and more assertive crossing motions compared with the observation trajectories.

### 5.5. Safety analysis using generated data

Real-world vehicle–pedestrian near-misses are rare. Argoverse 2 provides enough data for model training but not for large-scale safety analysis. It contains 411 Pedestrian-AV interactions (8,865 time steps) and 2,404 Pedestrian-HDV interactions (52,895 time steps). To address this limitation, this study trained two pedestrian policies (AV-based and HDV-based SMamba-DDPG) and used them to generate expanded datasets that capture vehicle-type-specific pedestrian responses.

This study first extracted AV and HDV trajectories at urban intersections from Argoverse 2 to seed data generation. We paired the AV-trained pedestrian model with AV trajectories and the HDV-trained model with HDV trajectories. Each pedestrian agent generated collision avoidance behavior based on its learned policy. This process created two large datasets of safety-critical interactions for analysis.

Each generated interaction captures vehicle–pedestrian temporal dynamics. At each time step $t$, the pedestrian state is:

$$S(t) = \left(\Delta D_{\mathrm{lon}}(t), V_{\mathrm{lon}}^{\mathrm{ped}}(t), \Delta V_{\mathrm{lon}}(t), \Delta D_{\mathrm{lat}}(t), V_{\mathrm{lat}}^{\mathrm{ped}}(t), \Delta V_{\mathrm{lat}}(t)\right) \tag{36}$$

Here, $\Delta D_{\mathrm{lon}}$ and $\Delta D_{\mathrm{lat}}$ are the longitudinal and lateral distances between pedestrian and vehicle. $V_{\mathrm{lon}}^{\mathrm{ped}}$ and $V_{\mathrm{lat}}^{\mathrm{ped}}$ are the pedestrian's velocities. $\Delta V_{\mathrm{lon}}$ and $\Delta V_{\mathrm{lat}}$ are the relative velocities. Each interaction is recorded at 0.1 second intervals. This study extracted sequences in which CurvTTC remained below 5 s throughout the interaction, indicating sustained collision risk. A total of 29,934 safety-critical sequences were identified for Pedestrian-AV interactions, with an equal number for Pedestrian-HDV interactions.

#### 5.5.1 Conflict risk assessment

The critical onset is defined as the first time step with CurvTTC < 10 s, marking entry into the collision-risk region. This research recorded the vehicle and pedestrian speeds at this onset frame as the initial speeds and flagged a sequence as a conflict if any frame had CurvTTC < 2 s, indicating imminent collision risk. Initial vehicle and pedestrian speeds were binned from 0–4 m/s in 0.5 m/s steps. For each bin, the conflict rate is calculated as:

$$\text{Conflict Rate } = \frac{\text{Number of sequences with a conflict}}{\text{Total number of sequences in the bin}} \times 100\% \tag{37}$$

For Pedestrian-AV interactions, conflict rates generally increase with both vehicle and pedestrian speed. When both agents move below 1.5 m/s, conflict rates remain minimal at 18-30%. The rates rise progressively with speed and reach 77% when both agents exceed 3.5 m/s. Figure 14 (b) reveals that most interactions occur at moderate speeds. The highest sample count (1,780) appears in the bin where vehicle speed is 2.5–3.0 m/s and pedestrian speed is 2.0–2.5 m/s. This region shows a moderate conflict risk of 40-44%. The 3D surface in Figure 14 (c) shows a monotonic risk increase across the speed grid. For Pedestrian-HDV interactions, conflict risk also increases with speed, peaking at 85% when both agents exceed 3.5 m/s, compared with 77% for AVs. Figure 14 (e) confirms that most samples are concentrated in the 2.0–3.0 m/s range. The 3D surface in Figure 14 (f) displays a steeper risk gradient, indicating greater danger in Pedestrian-HDV encounters at high speeds.

The comparison reveals distinct risk patterns for different vehicle types. Pedestrian-AV interactions consistently yield lower conflict rates than HDV interactions under identical speed conditions. This difference becomes most apparent at high speeds, where AV conflict rates are 10-15 percentage points lower than HDV rates. This pattern indicates vehicle-type-dependent differences in pedestrian avoidance kinematics under matched speed conditions.

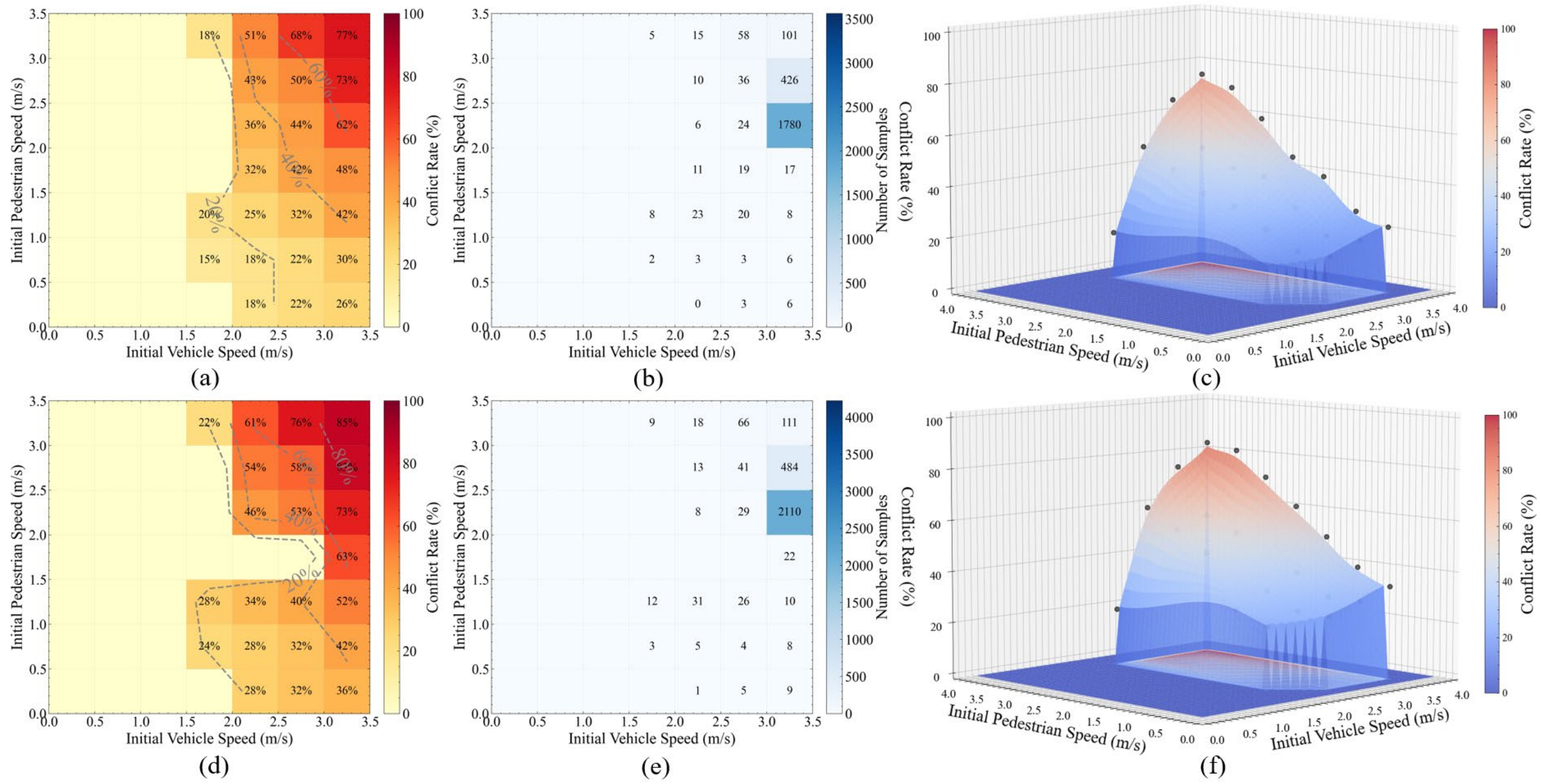

**Figure 14: (a)-(c) Conflict analysis for Pedestrian-AV interactions, showing rate heatmap, sample distribution, and 3D surface, respectively; (d)-(f) Corresponding analysis for Pedestrian-HDV interactions**

5.5.2 Yielding behavior

This study examined collision avoidance strategies in safety-critical events using datasets generated by the SMamba-DDPG framework. A total of 29,934 safety-critical sequences for Pedestrian-AV and Pedestrian-HDV cases were analyzed to identify avoidance patterns.

Two avoidance strategies were identified: Vehicle-Yield and Pedestrian-Yield. For each sequence, the critical onset was defined as the first frame with CurvTTC $< 10$ s, indicating entry into a collision-risk zone.

- Vehicle-Yield: assigned when the vehicle decelerated while the pedestrian maintained forward motion. The criteria were: vehicle deceleration $\leq -2.5$ m/s² or relative speed reduction $\geq 35\%$; pedestrian speed change $> -0.3$ m/s; and path length $\geq 3$ m, indicating continuous movement.

- Pedestrian-Yield: assigned when the pedestrian slowed or stepped back, while the vehicle proceeded with minimal braking. The criteria were: pedestrian speed reduction $\leq -0.5$ m/s or backward displacement $\geq 0.3$ m; vehicle deceleration $> -1.5$ m/s² with relative speed reduction $<$ 20%; and path length $\geq 8$ m.

All automatic labels were manually checked to ensure accuracy and consistency.

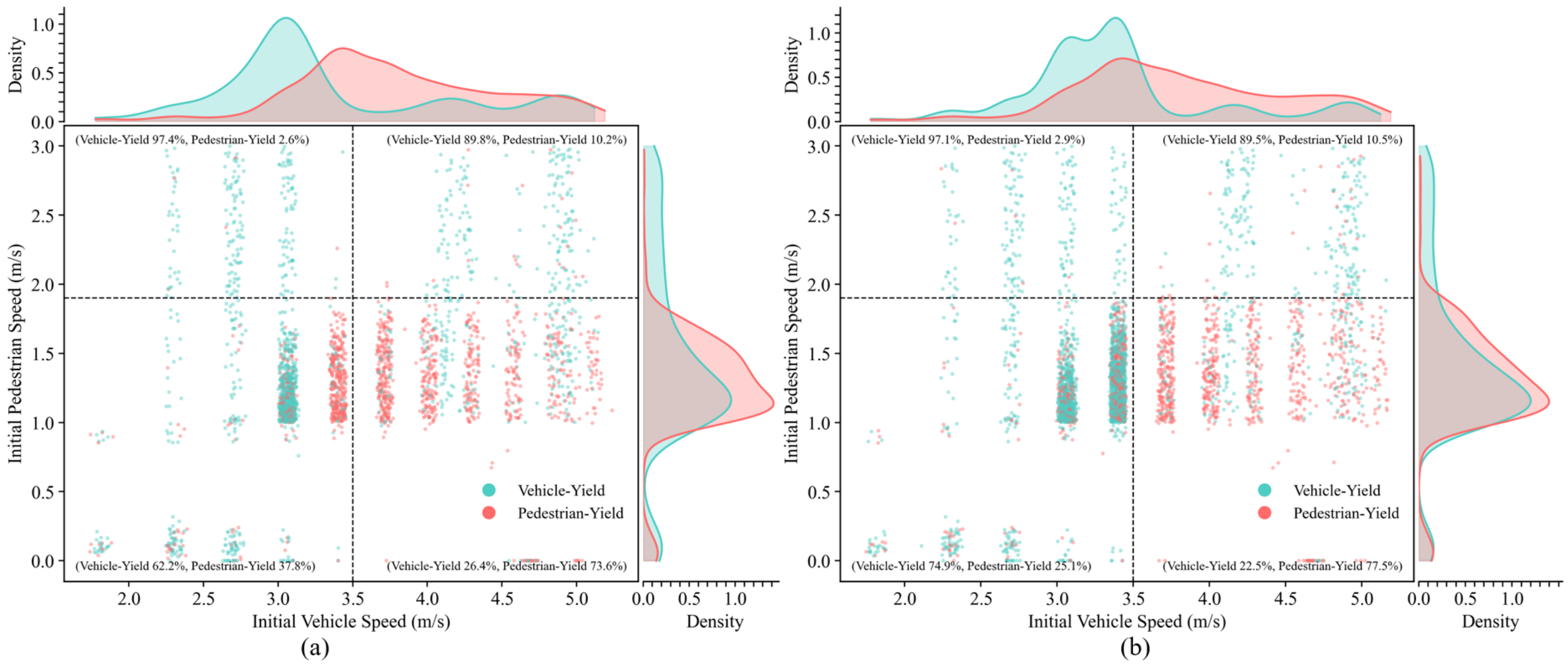

**Figure 15: Quadrant Analysis of Yielding Strategies by Initial Speed: (a) Pedestrian-AV Interactions; (b) Pedestrian-HDV Interactions**

To enable fair comparison between vehicle types, k-means clustering analysis was applied to the initial speed distributions, yielding unified thresholds of 3.5 m/s for vehicles and 1.9 m/s for pedestrians. These thresholds were applied uniformly to categorize yielding behavior in both Pedestrian-AV and Pedestrian-HDV interactions. Figure 15 (a) presents Pedestrian-AV interactions. The density plots along the top and right axes indicate that most interactions occur when pedestrians move faster than vehicles (upper-left region), where Vehicle-Yield dominates (97.4%). When AVs travel faster and pedestrians are slower (lower-right quadrant), Pedestrian-Yield becomes the primary response (73.6%). Even at low speeds for both agents (lower-left quadrant), AVs tend to yield (62.2%), reflecting their conservative control strategies. Figure 15 (b) shows Pedestrian-HDV interactions using the same speed thresholds. HDV interactions cluster more toward mid-to-high vehicle speeds, with a larger share of Pedestrian-Yield events in the lower-right quadrant (77.5%). While Vehicle-Yield remains dominant in the upper-left quadrant (97.1%), the lower-right quadrant shows a higher Pedestrian-Yield rate (77.5%).

Overall yielding patterns reveal distinct behavioral differences between vehicle types. Pedestrian-Yield accounted for 46.07% of all avoidance events in Pedestrian-AV interactions, compared with 37.15% in Pedestrian-HDV interactions. This higher pedestrian yielding rate in AV scenarios indicates more cautious and conservative crossing behavior, suggesting that pedestrians exhibit greater vigilance toward automated vehicles, which may reflect unfamiliarity with their decision-making patterns or uncertainty about their responsiveness in dynamic traffic situations.

## 6. Conclusion

This study developed a novel framework to model and quantify pedestrian avoidance behavior differences between AV and HDV encounters in safety-critical conditions. The SMamba-DDPG framework learns distinct pedestrian policies from high-risk interactions in the Argoverse 2 dataset. Separate models were trained for each vehicle type, enabling direct behavioral comparison, counterfactual analysis, and large-scale safety assessment.

The findings quantify significant differences in pedestrian behavior across vehicle types. The velocity-based SMamba-DDPG variant ($\mathrm{SMamba} - \mathrm{DDPG}_v$) achieved the best trajectory reconstruction accuracy across both scenarios, with ADE of 0.65 m, and FDE of 1.12 m for Pedestrian-AV interactions, and ADE of 0.64 m, and FDE of 1.13 m for Pedestrian-HDV interactions. These results indicate performance improvements over both reinforcement learning–based and supervised learning models. Reconstruction validation demonstrated that SMamba-DDPG accurately reconstructs pedestrian crash avoidance behaviors in both AV and HDV scenarios, with low RMSE confirming the model's effectiveness. Reaction time analysis validated the model's effectiveness in reproducing human-like response delays, showing no significant differences between reconstructed and observed data. Pedestrians exhibited lower reaction times when responding to AVs compared to HDVs. Counterfactual analysis showed that pedestrian crossing behavior differed between vehicle types, with pedestrians exhibiting slower crossing speeds when interacting with AVs compared to HDVs. Large-scale safety analysis of model-generated data showed that Pedestrian-AV interactions consistently yielded lower conflict rates than Pedestrian-HDV interactions across speed conditions. Analysis of yielding strategies showed pedestrians exhibited more cautious crossing behavior toward AVs, with a higher pedestrian yielding rate (46.07%) compared to HDV interactions (37.15%).

This research provides methodological, empirical, and analytical contributions to the literature. Methodologically, it presents a robust SMamba-DDPG framework for modeling the abrupt, sequential dynamics of pedestrian behavior in safety-critical events. Empirically, the study provides quantitative evidence of distinct pedestrian responses to AVs and HDVs, highlighting the limitations of uniform modeling in mixed-traffic environments. Analytically, the study uses counterfactual simulations and conflict analysis to provide insights into how vehicle automation affects pedestrian risk perception, decision-making, and yielding behavior.

In summary, the study emphasizes the need for vehicle-type-specific behavioral models to enhance the realism of traffic simulations and the effectiveness of AV safety systems. The proposed framework enables realistic pedestrian-response generation, supporting the design of AVs that interact safely and intuitively with vulnerable road users. Future work should expand the dataset to encompass more diverse traffic environments to improve model generalizability. Additionally, exploring more sophisticated reward structures may enhance the model's ability to capture human decision-making dynamics.

**Acknowledgment**

The contents of this paper present the views of the authors, who are responsible for the facts and accuracy of the data presented herein. The contents of the paper do not reflect the official views or policies of the agencies. The work was supported by the National Science Foundation under Grant Nos. 2315450 and 2315451 and by the Transportation Informatics Lab, Department of Civil & Environmental Engineering at Old Dominion University (ODU).